\documentclass[lettersize,journal]{IEEEtran}
\usepackage{amsmath,amsfonts}

\usepackage{algorithm}
\usepackage{algpseudocode}

\usepackage{array}
\usepackage[caption=false,font=normalsize,labelfont=sf,textfont=sf]{subfig}
\usepackage{textcomp}
\usepackage{stfloats}
\usepackage{url}
\usepackage{verbatim}
\usepackage{graphicx}
\usepackage{cite}
\begin{document}

\title{Adaptive Guided Upsampling for Low-light Image Enhancement}

\author{Angela Vivian Dcosta, Chunbo Song, Rafael Radkowski
        % <-this % stops a space
\thanks{Lenovo Research, 7001 Development Drive, Morrisville, NC 27560}% <-this % stops a space
\thanks{Manuscript received October 3, 2025}}

\markboth{October~2025}%
{Adaptive Guided Upsampling for Low-light Image Enhancement}


\maketitle

\begin{abstract}
We introduce Adaptive Guided Upsampling (AGU), an efficient method for upscaling low-light images capable of optimizing multiple image quality characteristics at the same time, such as reducing noise and increasing sharpness. It is based on a guided image method, which transfers image characteristics from a guidance image to the target image. Using state-of-the-art guided methods, low-light images lack sufficient characteristics for this purpose due to their high noise level and low brightness, rendering suboptimal/not significantly improved images in the process. 
We solve this problem with multi-parameter optimization, learning the association between multiple low-light and bright image characteristics. 
Our proposed machine learning method learns these characteristics from a few sample images-pairs. AGU can render high-quality images in real time using low-quality, low-resolution input; our experiments demonstrate that it is superior to state-of-the-art methods in the addressed low-light use case.
\end{abstract}


\section{Introduction}
The enhancement of low-light images is gaining relevance, especially for video conferencing applications that support the hybrid work culture. Images and videos captured under these conditions ($\leq$20 lux) suffer from several degradations, such as under-saturation \cite{Wang:2022a}, loss of texture/blur, and high noise \cite{Kumar:2016}, to mention a few \cite{Zhang:2021}. These degradations reduce the image quality compared to the user's experience under regular-light conditions. Low-light enhancement methods encompasses various techniques and algorithms aiming to mitigate different negative properties of low light captures such as high noise, low sharpness, and saturation, to the point that the processes image meets its regular light (200lx) equivalent. In this context, low-light image enhancement refers to techniques and methods that improve the brightness/contrast/saturation/sharpness, and the noise of images captured in environments with reduced light. An enhanced low-light image (also referred to as a high-quality image) should be comparable to a bright, 100 lux equivalent image, showing saturated colors, be sharp, and contain no visible noise. Although several low-light image enhancement methods exist (\cite{Wang:2023a}, \cite{Lu:2023}, \cite{Zhang:2023a}, \cite{Park:2024}), our primary challenge is to increase sharpness and reduce noise after brightening an image.

Finding a general solution for noise reduction and sharpness improvement is still an ongoing research topic, with the relevant literature suggesting various linear and nonlinear techniques. Classically, the bilateral filter \cite{Tomasi:1998, Aleksic:2006} is likely one of the most renowned techniques. It is a nonlinear, edge-preserving filter that relies on averaging image content to reduce noise. Spatial and frequency domain filters  \cite{Gonzalez:2017}, such as mean, median, and notch filters, rely on noise models or assumptions made about the noise characteristics \cite{Kumar:2020, Jasim:2019}. Although the noise reduction performance is appropriate in certain conditions, our observations show that they often over-smooth the image, resulting in loss of textures, details, and sharpness in general \cite{Fan:2019, Verma:2013}.

Recent approaches, such as content- or guided-filters and neural networks, offer a more fine-tuned control of noise reduction, preserving edges and sharpness for a vast range of capture conditions. In particular, guided filters suggest solutions that reduce noise and maintain or increase sharpness. They rely on the idea of a guiding image (\cite{PJ15}, \cite{Kim:2004}): transfer sharpness and noise characteristics from the guidance image to the image to be improved (target image) using a linear transform. Compared to the original bilateral filter \cite{Tomasi:1998} and similar, these filters outperform classical approaches by avoiding assumptions \& adaptive measures. Instead, they rely on machine learning techniques to estimate appropriate corrections for various conditions. However, the linear transform capabilities are limited in scope, which limits their performance in the process. They especially lack the means to transfer image characteristics from a low-light guidance image, which is critical for low-light image enhancement where the brightness of the guidance image (low-light) and target image (increased brightness) differ, yielding diminishing noise reduction and sharpening performance.

On the other hand, deep neural networks are more versatile and can account for various image attributes, including brightness, noise, and sharpness. Although autoencoders and other generative models’ results are comparable to classical filters, \cite{Vincent:2008, Kumar:2020}, newer generative models outperform those. It is likely that modern deep learning approaches such as Denoise GANs (\cite{WG21, LN20}) can overcome the challenges of adaptive and smoothing filters. However, generative models currently fail to meet our real-time video frame (15fps low-light, 30fps otherwise) and power ($\leq$3 watts) requirements for implementations on current GPUs and NPUs. We require a lightweight solution to meet camera runtime requirements without compromising image quality.

We propose Adaptive Guided Upsampling (AGU), a method that employs low-light guidance images to transfer high-quality image characteristics into an  image of increased size (also referred to as upsampled image). It learns the association between low-light and bright image characteristics and applies learned parameters to the target image during inference. Although the primary function of AGU is to upsample an image, its main challenge is to perform this task while maintaining sharpness and reducing noise. We drew motivation from the Adaptive Guided Filter (AGF, \cite{PJ15}) and Fast Guided Filter (FGF, \cite{HS15}); however, both methods underperform when working with low-light images (see results in Section 5). Our use case is low light image enhancement for camera applications such as video conferences, footage recording in low-light situation, among others, where users intend to record with the best camera quality they can obtain from their laptop camera. To facilitate this scenarios, we contribute
%To overcome this problem, we contribute
\begin{enumerate}
\item a multi-parameter model that characterizes differences between image characteristics (brightness, sharpness, noise) of low-light and high-quality images.
\item a machine learning-based method to solve the multi-parameter optimization problem to preserve/restore image brightness, high-sharpness, and reduce noise in the output image.
\item results demonstrating the advantages of our method when working with low-light images. 
\end{enumerate}

The paper is structured as follows: the next section discusses the related work. Section \ref{agf} introduces the Adaptive Guided Filter (AGF) as relevant background information. Our approach is covered in Section \ref{sec:method}. Section \ref{sec:results} presents our experiments and results. We conclude with our findings in Section \ref{sec:conclusion}.

\section{Related Work} \label{sec:related_work}

Numerous methods for image upsampling, sharpness improvement, and noise reduction have been studied. In case of images, noise is an alteration or variance in the pixel content generated during the image acquisition process. It originates from various sources such as photon shot noise, gain noise, and color noise (or de-mosaic noise). All noise results in grainy visual artifacts and lowers perceived image quality. Sharpness, on the other hand, is defined by the contrast difference or rise difference of neighboring image pixels -- high sharpness results in a crisp image with defined edges and contours. This section covers multiple techniques from classical filters (i.e., \cite{Motwani:2004}, \cite{Tukey:1980}, \cite{Fan:2019}, \cite{Strela:2001}) to deep neural networks. We refer to surveys such as \cite{Fan:2019} and \cite{Rajshree:2017} for a complete summary.

\subsection{Guidance Filters}
Guidance or guided filters rely on the idea of a reference image - the guidance image - used to transfer characteristics from a prior to a target image. They aim for a best-fit relationship between the guided and filtered signal, often with the goal of reducing image noise while maintaining sharpness. He et al. (\cite{HS10}) demonstrated one of the first solutions using a local linear model to achieve the best fit. A later version used a box filter to speed up processing \cite{HS15}. Mishiba et al. \cite{Mishiba:2023} suggested a Fast Guided Median filter, which is based on He's approach but allows for more performance. Several other solutions have been presented since then such as \cite{PJ15}, \cite{LW2021}, \cite{CA16}, \cite{KMDM:2007}, \cite{ZA07}, \cite{Kim:2004}, \cite{Xu:2018a}, \cite{Wang:2017}.The authors of \cite{LW2021} discuss an improved adaptive spatial filter, which uses a linear combination of a range and average filter. Xu et al. (\cite{Xu:2018a}) suggest using a fuzzy clustering method to identify parameters guiding noise reduction. We share the same goal but use multi-step linear optimization to adjust the noise-reduction and sharpness parameters. Other guided filter approaches also perform upsampling, employing various means such as a tone mapping curve \cite{CA16} or sub-sampling information from a high-resolution image to upsample and enhance a low-resolution image \cite{KMDM:2007}, as well as combining the guided filter with other methods (\cite{Wang:2017}). 

Our solution was primarily inspired by Pham et al, \cite{PJ15}, the Adaptive Guided Filter (AGF. see Section \ref{agf}), which uses linear optimization (exhaustive search) to align the output image with the guidance image sharpness and noise levels. We also want to highlight Bilateral Guided Upsampling (BGU),\cite{CA16}, which learns the difference between a high- and low-resolution image using an affine transform to recover sharpness during upsampling. 

All the mentioned filters work as advertised when input and guidance images are of the same brightness or scale. However, they lack means accounting for variances of those attributes. Filters that perform upsampling can work on different scales but lack brightness invariance.

\subsection{Bilateral Filter}
The bilateral filter is a classical means to reduce noise while preserving edges (and thus sharpness) in an image \cite{Tomasi:1998, Weiss:2006}. It is a nonlinear, edge-preserving kernel filter \cite{Tomasi:1998, Weiss:2006} using weighted range and chromatic Gaussian filters to improve image content. Multiple variants have been introduced over the years, such as in \cite{Aleksic:2006}, \cite{KSSV:2019}, \cite{Kumar:2016}, \cite{KSSV:2019}, \cite{Kaur:2017}.

For instance, \cite{Aleksic:2006} suggests using a sharpening mask along with the bilateral filter to improve edge slopes. Other authors combined a bilateral filter with an unsharpen mask \cite{KSSV:2019}, \cite{KSSV:2019}, adaptive filters, \cite{Kumar:2016}, \cite{ZA07}, adaptive filters based on edge detection \cite{Kaur:2017} and  low-pass filters (\cite{Chen:2020a}) to adjust the spatial kernel range. Unwanted side effects include over-smoothed textures and artifacts such as staircase effects. Various remedies for over-smoothing exist, and iterating those exceeds the scope of this review. For example, \cite{Zhou:2022} suggests using a sliding window, a measure that moves the pixel of interest from the center of the kernel to the leading edge. 

Bilateral filters are known to be costly. Several authors address this problem using various methods such as parameter approximation based on subsampling \cite{Yang::2022} or polynomial approximation \cite{Gavaskar:2018} to reduce computational complexity. O(1) algorithms have been suggested \cite{Porikli:2008}, \cite{Tu:2016}, leveraging pre-processing (e.g., histograms) and look-up tables, among other means. Other approaches employ parallel computing \cite{Yang:2009} or hardware implementation with VLSI (\cite{Wu:2016}) to achieve real-time performance.

\subsection{Optimization}

Optimization and machine learning techniques generally are other alternatives for image denoising and upsampling. Techniques are mainly employed to identify an optimal parameter threshold, discriminate sets, and others. This section refers to work that is relevant to our effort since it also relies on optimizing parameters. Here, we address various optimization strategies for parameter identification beyond guided and bilateral filters. 

For instance, Vaiyapuri et al. \cite{Vaiyapuri:2021} estimated the optimal parameter threshold for a denoiser using a genetic algorithm. Other efforts focus on simulated annealing \cite{Knaus:2014}, constraint satisfaction \cite{Mahdaoui:2022}, \cite{Reyes:2013}, differential evolution \cite{Chandra:2014}, and sparse optimization \cite{Xu:2018b}. Efforts that are close to our approach have been introduced in \cite{Kim:2004}, \cite{Beck:2009}, and \cite{Mahdaoui:2022}. Kim et al. \cite{Kim:2004} used linear optimization to identify the best parameters for an unsharpen mask.  Beck et al. \cite{Beck:2009} linearly optimized the parameters for a total variation algorithm, and Mahdaoui et al. \cite{Mahdaoui:2022} optimized a linear compression model to minimize compression artifacts directly. Su et al. \cite{Su:2018} suggest a noise-aware filter estimating the noise based on the luminance change in an image. Li et al. \cite{Li:2020} describes a low-light enhancer based on a complementary gamma function; however, the method's noise and sharpness performance is unclear. All approaches use linear optimization with a different goal. We use a similar optimization method but aim for extended guided filter parameters.

\subsection{Deep Learning}
Deep learning and all variations of convolutional neural networks also offer capable noise reduction solutions while maintaining sharpness simultaneously. They likely pose an alternative solution to our problem at hand. Various approaches have been introduced starting with denoise auto-encoders (\cite{VL10}), to multiple supervised convolutional architectures (\cite{ZZ+:2017}, \cite{ZZ+:2018}, \cite{MS+:2016}, \cite{TY+:2017}, \cite{Soh:2021}, \cite{ZZ+:2018}), \cite{Jiang:2021}, and others (\cite{Harvey:2022}). Promising approaches for our problem are diffusion models (\cite{Choi:2021}, \cite{Song:2021}, \cite{Saharia:2021}, \cite{Kawar:2022}, \cite{Song:2020b} ) and GANs (\cite{Chai:2019}, \cite{Kim:2019},  \cite{Linh:2020}).

Diffusion models learn a continuous Markov process to generate a probable dataset distribution. Ho et al. (\cite{Ho:2020}), for instance, suggested a denoising diffusion model based on non-equilibrium thermodynamics theory. The diffusion process adopts Langevin dynamics to generate the output distributions. The authors of \cite{Nichol:2021} extended this process to render it more controllable. Shi et al. (\cite{Shi:2022}) investigate a denoising model using a Schr{\"o}dinger bridge to describe the diffusion process; it incrementally reduces the KL-divergence between the data and target distribution. Karras et al., (\cite{Karras:2022}) describes various model and process changes for diffusion models to improve image quality. Yang et al. (\cite{Yang:2023}) present improved sampling methods that allow one to control the noise in the target distribution better. All approaches yield sufficient noise reduction and sharpness in comparison to classical techniques.

GANs are the most promising alternative compared to guided filters. To highlight selected examples, the authors of \cite{Chai:2019} suggest a two-way (cyclic) GAN to improve colors in an image, comparing the original input to the reversed, generated image. One strategy for noise reduction is to generate better input data for training; \cite{Kim:2019} follows this approach. They also investigate the effect of residual blocks on the noise reduction performance. The authors of \cite{Linh:2020} propose a similar two-step (network) GAN solution. The first network estimates the image noise, while the second one is trained on synthetic images using the noise estimates. Besides training with synthetic images, other authors work with GANs to generate noise-free images relying on conventional training data, e.g., in \cite{Chen:2020b}. Wang et al. (\cite{Wang:2022b}) improves edge preservation while denoising. The authors identify edges in frequency space and prevent blurring them. \cite{Chen:2018} describes the combination of noise reduction and super-resolution by introducing shared building blocks in a joint network architecture. Wang et al., \cite{Wang:2020a} uses a deep residual network as the generator GAN. Various other approaches have been suggested. Listing all of them goes beyond the scope of this paper; we refer to Dey et al., \cite{Dey:2020} who published a survey.

Besides GANs and diffusion models, several authors introduce convolutional architectures of various configurations to solve the denoising problem. For instance, Ran et al. \cite{Ran:2021} suggests a task-driven approach for training that relies on images grouped based on their content similarity. This approach increases image utilization. Guo et al. \cite{Guo:2020} suggest a two-step approach similar to ours. Their first step enhances low-light images, whereas the second step uses a neural network for denoising. Various other methods for convolutional denoising exist, such as self-supervised methods in \cite{LM+:2018}, \cite{KB+:2018}, \cite{WL+:2020}, \cite{BR:2019},\cite{TS+:2021}, \cite{MS+:2020}, \cite{XH+:2020}, \cite{PZ+:2021}, \cite{ZZ+:2017}, \cite{Ning:2021} and others \cite{SKP2020}, \cite{KAVK2006}, \cite{X2011}.

Despite their demonstrated effectiveness, neural network processing resource requirements are high, especially for image processing tasks. Although modern NPU processors (Intel Meteor Lake, AMD Phoenix, and beyond) are available that specialize in neural processing, frame rates and power consumption requirements still mandate being selective with solutions running on those processors. Thus, we opted for a more conventional denoising approach.

\subsection{Super-Resolution}
Single Image Super-Resolution (SISR) seeks to reconstruct a high-resolution (HR) image from its low-resolution (LR) counterpart. This task has been significantly advanced by deep learning, particularly convolutional neural networks (CNNs), attention mechanisms, and generative models.
Early deep learning approaches, such as SRCNN \cite{DCK:2014} and VDSR \cite{KKL:2015}, demonstrated the potential of CNNs for super-resolution by directly learning the mapping between LR and HR images. These were followed by deeper and more effective architectures like EDSR \cite{LSK:2017}, which improved performance by removing normalization layers, and RCAN \cite{ZLL+:2018}, which introduced channel attention to better capture informative features.

To address the limitations of local receptive fields in CNNs, attention-based and Transformer-inspired models have gained traction. SwinIR \cite{LCS:2021} employs a hierarchical Swin Transformer with shifted window attention, achieving a strong balance between accuracy and efficiency. More recently, HAT (Hybrid Attention Transformer, \cite{CWZ+:2022}) has integrated both channel and spatial attention, setting new benchmarks for image restoration quality.

In addition to deterministic models, generative approaches have shown strong performance in enhancing perceptual quality. SRGAN \cite{LTH+:2016} introduced adversarial training for photo-realistic detail generation, which was later improved by ESRGAN \cite{WYW+:2018} through Residual-in-Residual Dense Blocks and better perceptual loss functions. Real-ESRGAN \cite{WXD:2021} extended this work to real-world scenarios, training on more realistic degradations and using a U-Net discriminator for enhanced robustness.

A more recent direction involves diffusion-based methods. Models such as SR3 \cite{SHC+:2021} and IR-SDE \cite{LGZ+:2023} leverage denoising diffusion processes to progressively generate high-quality images from random noise. These models achieve state-of-the-art perceptual results, albeit with significantly higher computational costs.

\section{Adaptive Guided Filter} \label{agf}
%\subsection{Adaptive Guided Filter}
The Adaptive Guided Filter (AGF, \cite{PJ15}) is a linear transform filter that uses a guidance image $G$ to control the noise reduction and sharpness increase of an input image $I$. The filter is motivated by the Guided Filter \cite{HS10}, adopting the main idea of leveraging image content to define the parameters for noise-reduction and sharpening. Simplified, one must denoise/blur image content in flat areas and sharpen along its edges. AGF transfers those image characteristics from $G$ to $I$ using optimization. 

AGF works with the linear transformation as follows:
\begin{equation}  \label{eq:01}
{O_p} = A_{k}(G_{p}+\xi_{p})+B_{k}
\end{equation}

\noindent with $A_k$ and $B_k$, the linear transform coefficients yielding an output image ${O_p}$ from $G$ with image characteristics similar to the guidance image $G$, and $\xi_p$, a trainable sharpness correction parameter; with $p$, per pixel where $p\in w_{k}$, and $k$, applied per kernel. To solve Eq~\ref{eq:01} for the coefficients $A_k$ and $B_k$, the authors \cite{PJ15} suggest using the closed-form solution:
\begin{align} \label{eq:calc_ab}
    A_{k} &= \frac{(1/|w|)\sum_{p\in w_{k}}^{}(G_{p}I_{p}-\bar{G_{k}}\bar{I_{k}})}{{\sigma}_{k}^{2}+\epsilon}\\
    B_{k} &= \bar{I_{k}}-A_{k}\bar{G_{k}}
\end{align}

Solving for $A_k$ and $B_k$ is nested into an optimization task identifying parameters for $\xi$ and $\varepsilon_k$ using the cost function \cite{PJ15}:
%To solve for $\xi$ linear transform, \cite{PJ15} suggests a linear regression solution using the cost function:
\begin{equation} \label{eq:02}
    E(A_{k},B_{k}) = \sum_{p\in w_{k}}^{}((A_{k}G_{p}+B_{k}-I_{p})^{2}+\epsilon A_{k}^{2})
\end{equation}
\noindent The authors optimize the parameters using exhaustive search \cite{Kim:2004}.

The two parameters, $\epsilon$ and $\xi$, facilitate control over sharpness improvement and noise reduction. The regularization parameter $\epsilon$ acts as a smoothing factor for noise, allowing refined gradient-based noise reduction. It is calculated as $\epsilon = \lambda \sigma^2$, with $\lambda$, a weight parameter, and $\sigma$, a trainable optimal variance. The parameter $\xi_p$ in Eq.~\ref{eq:01} controls sharpness with the goal of maximizing it along edges and maintaining minimal noise in uniform areas.  

To achieve an optimal output image, $\xi$ and $\epsilon$ require differential behavior for uniform areas vs. edges. Therefore, the authors adopt a class-based sharpening method that discriminates edges from areas by assigning classes to edges depending on the edge magnitude. They leverage a Laplacian of Gaussian (LoG) impulse response to gradually discriminate between homogeneous areas and strong edges. The LoG response is given by:
\begin{equation}\label{Log_kernel}
LoG(x,y) = -\frac{1}{\pi\sigma^4}(1-\frac{x^2+y^2}{2\sigma^2})e^{\frac{-(x^2+y^2)}{2\sigma^2}}
\end{equation}
\noindent with $p=(x,y)$. The LoG-response ranges between min-max bounds, discretized into classes, and values for $\xi$ and $\epsilon$ are identified per class. 

The parameter $\xi$ applied can yield intensity bounds beyond the guidance image $G$. To prevent this, it is limited by:
\begin{equation}\label{eta_bounds}
\xi_{p}^{'}=\left\{ \begin{array}{cc} 
MAX(G,w_{k})-G_{p},& \text{if } G_{p}+\xi^{*}_{p} > MAX(G,w_{k})\\ 
MIN(G,w_{k})-G_{p},& \text{if } G_{p}+\xi^{*}_{p} > MIN(G,w_{k})\\ 
\xi^{*}_{p}& \text{otherwise }\\
\end{array} \right.
\end{equation}
\noindent The equation corrects the sum of any correction from $\xi$ and the image $G$ (kernel-wise) to the maximum of the kernel $w_k$.

% solving
\noindent We refer to \cite{PJ15} and \cite{Kim:2004} for the full solution.

During inference, one solves Eq.~\ref{eq:01} and \ref{eq:calc_ab}. The inference algorithm selects the parameters $\xi$ and $\sigma$ per pixel based on their class label. A pixel-wise LoG is computed for each image to identify the class. The output is an image with reduced noise in homogeneous areas but sharpened edges. 

Our results for AGF (see Section~\ref{sec:results}) suggest sharpness improvements compared to a bilateral filter when applied in our low-light use case. The results demonstrate that AGF is limited to images of similar brightness and scale for $I$ and $G$. This is no surprise since the authors of \cite{PJ15} focused their work on the impact of noise reduction and simultaneous sharpening, experimenting primarily with artificial noise. We notice a discrepancy when exposing the model to our typical laptop camera and low-light images, where $\xi$ fails to train correctly. 
\section{Adaptive Guided Upsampling Method}\label{sec:method}

Adaptive Guided Upsampling (AGU) converts the image enhancement problem into a multi-parameter optimization problem, identifying parameters to enhance an image for optimal brightness, noise, and sharpness. AGU focuses on low-light images, employing them as guidance images (referred to as low-light guidance images or just guidance images). It is brightness-agnostic, referring to the capability of using input and guidance images of different brightness levels. It is also scale-agnostic since input, output, and guidance image can be of different scales. Compared to the previous art, we introduce a novel method to optimize relevant model parameters (see Section~\ref{agf}) and an extended model supporting additional optimization targets.

Sections~\ref{sec:brightness} and \ref{sec:scale} explain the model and its linear transformation properties. Section~\ref{sec:train} describes our training strategy. The following section starts with an overview. 

\subsection{Overview and Application} \label{sec:overview}
\begin{figure*}[h]
\begin{center}
\includegraphics[width = 1.6\columnwidth]{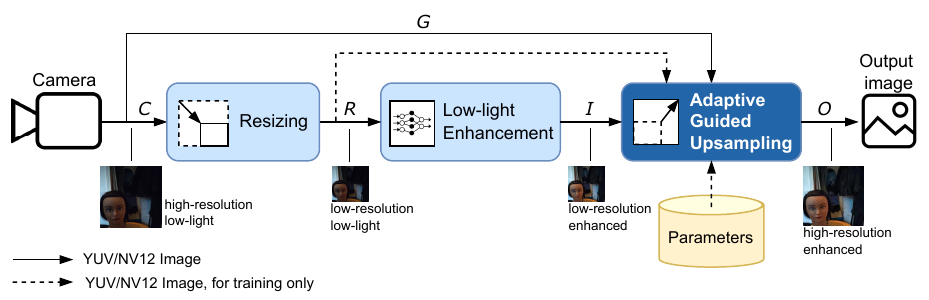}
\caption{Processing pipeline illustrating our application example, with $C$ and $G$, the camera/guidance image. $R$, a resized, low-resolution image, $I$, the input image for Adaptive Guided Upsampling, and $O$, the output image. Note that $I$ is a brightness and color-enhanced low-resolution image, where $O$ is the final high-resolution result. AGU maintains sharpness of the original image along edges and reduces noise in uniform areas.  }
\label{fig:Figure1}
\end{center}
\end{figure*}

Figure~\ref{fig:Figure1} illustrates the functional processing pipeline of our application example, where AGU is the relevant component for this contribution. In our low-light use case, all images with a brightness of $\leq 20 lux$ are considered low-light images. %The goal of this low-light enhancement objective is a brighter image, increasing the brightness from low-light to a $100 lux$ equivalent while reducing noise and increasing sharpness. 
The output and goal of the our method is a brighter image, increasing the brightness from low-light to a $100 lux$ equivalent while reducing noise and increasing sharpness in comparison to the input image. 

% Note, maintaining sharpness is an fuzzy goal for low-light images since they camera image captured at low-light is already of reduce sharpness. 

% Figure needs to go here

Input to the process is the low-light image $C$ (usually from a camera), also acting as the guidance image $G$. It is down-scaled in the first step for performance reasons. Full-resolution images (usually 1080p-2160p) exceed our compute resource target when using a neural network, including the one we employ (see \cite{FSM+22} for details). Although cameras operating in low-light conditions reduce the frame rate to 10-15 fps, the extended processing time saturates available resources easily. We resize the image to a lower resolution image $R$ (e.g., 540x960) to maintain camera frame time and remain within power-consumption limits. 

Subsequently, the low-light enhancer improves the image $R$. We use a UNet-based convolutional neural network processing the images in YUV format \cite{FSM+22}. The result is a enhanced low-light image, a brightness-enhanced image $I$ with improved colors and contrast in low resolution, referred to as the "enhanced image" in the following text. Although the low-light enhancer improves colors and brightness, it retains noise. Noise in a brighter image appears emphasized and more saturated, which is an unintended and unwanted side-effect.  

%%%%%
Next, we employ our method to remove noise and to upscale the image to the original camera resolution. Like AGF, we use a linear transform, which our model extends and solves in a multi-parameter optimization fashion: 

\begin{equation}
O^{\uparrow}_p = \mathcal{I}(A_k)  (G_p + \tau_p + \xi_p)  + \mathcal{I}(B_k), \\
\end{equation}
with the linear coefficients $A$ and $B$, encoding the characteristics of the input image $I$. Identifying $A$ and $B$ requires an input image $I$. We use the low-res, low-light camera image for this purpose. With $G$ as the guide, the linear transform performs characteristics transfer from $G$ to $I$, including noise reduction and sharpening. Noise reduction, as described in Section~\ref{agf}, is achieved by using a regularization during training of $A$ and $B$, which ensures a smooth image in uniform areas and sharp edges by employing class-based correction. Here, $\xi$ remains the class-based sharpening factor.  The coefficient $\tau$ is a brightness adjustment factor we introduced to correct for brightness differences between the guidance and input images. All coefficients are either applied per pixel $p$ or kernel $k$. The function $\mathcal{I}$ is our interpolation function for image resizing. Although low-light, with some contrast adjustments, it maintains the maximum number of details and sharpness. The output is an upscaled, low-noise, sharp image $O^{\uparrow}$. 

Our model addresses two challenges: First, we account for brightness difference in $G$ and $I$, since they are of different brightness levels. Our experiments with the AGF (Section~\ref{agf}) demonstrate its performance declines in the low-light use case, since the model parameters optimize to adjust brightness.
Secondly, we address the resolution difference between the input image $I$ and the guidance image $G$ during upscaling. This difference effects sharpness restoration since features in the low-res image are under-emphasized and lack robustness to restore sharpness in a high-res image properly. To account for both challenges, we contribute an extended method and also adapted the training process (see Section~\ref{sec:results} for training), working with low-light content and images of different scales.

\subsection{Brightness-Agnostic Guidance Image} \label{sec:brightness}
%Brightness agnostic accounts for the brightness difference between the training image $I$ and guidance image $G$. In our use case, the image to be upscaled is typically of higher brightness than the guidance/camera image $G$. Figure~\ref{fig:offset} illustrates the problem and the effect of our solution. The figure shows a) a guidance image $G$ in regular light, b) a low-light image $G$, and c) an input image $I$ with reduced sharpness. The brightness values have been sampled along an edge (red line sampled across 1-2). 

Brightness agnostic accounts for the brightness difference between the training image $I$ and guidance image $G$. In our use case, the image to be upscaled $I$ is typically of higher brightness, but lower contrast, than the guidance/camera image $G$. If unaddressed, the parameter $\xi$ trains to adjust the brightness difference between the two images, but it neglects the contrast reflecting sharpness in the image. Brightness differences are of higher magnitude and allow for a faster loss reduction than low-magnitude sharpness differences do. 

The following narrative first explains the problem in detail before we introduce our approach. 

%%---------------------------------------------------------------------------------------

\subsubsection{Problem}
%% New paragraphs describing the problem. 
The primary problem with the parameter $\xi$ trained to adjust brightness originates in the training process using gradient descent. Considering the loss function as the starting point: 

\begin{equation}
E(A_k, B_k) = \sum_{p\in w_{k}}\left( (A_k (G_k+\xi_p) + B_k - I_k)^2 + \epsilon A_k^2 \right)
\end{equation}

\noindent with $O = A_k (G_k+\xi) + B_k$; note that the equation and the following narrative in this section is void of any scale nomenclature to simplify reading. The first derivative of $E$ solves to:
\begin{equation}
\frac{\text{d}E}{ \text{d}\xi} = 2(O_k-I_k)A_k
\end{equation}

One can see that the gradient $\text{d}E$ is of high magnitude when the difference between $|O|$ and $|I|$ is high, corresponding to images with different brightness, consequently with sharpness($I$) > sharpness($O$) due to the brightness difference. If $|O| \approx |I| \implies E \rightarrow 0$; if $|O| >> |I| \implies E \rightarrow \max$ ($255^2$ in this case). With a higher error $E$, gradient descent has a higher incentive to develop $\xi$ towards $|O| == |I|$ from $|O|>>|I|$ as a starting point instead of adjusting the more subtle contrast magnitude peaks along edges that account for sharpness. The trained parameters of $\xi$ are too high in this case to create sharp edges.

\begin{figure}[h!]
\centering
\includegraphics[width =\columnwidth]{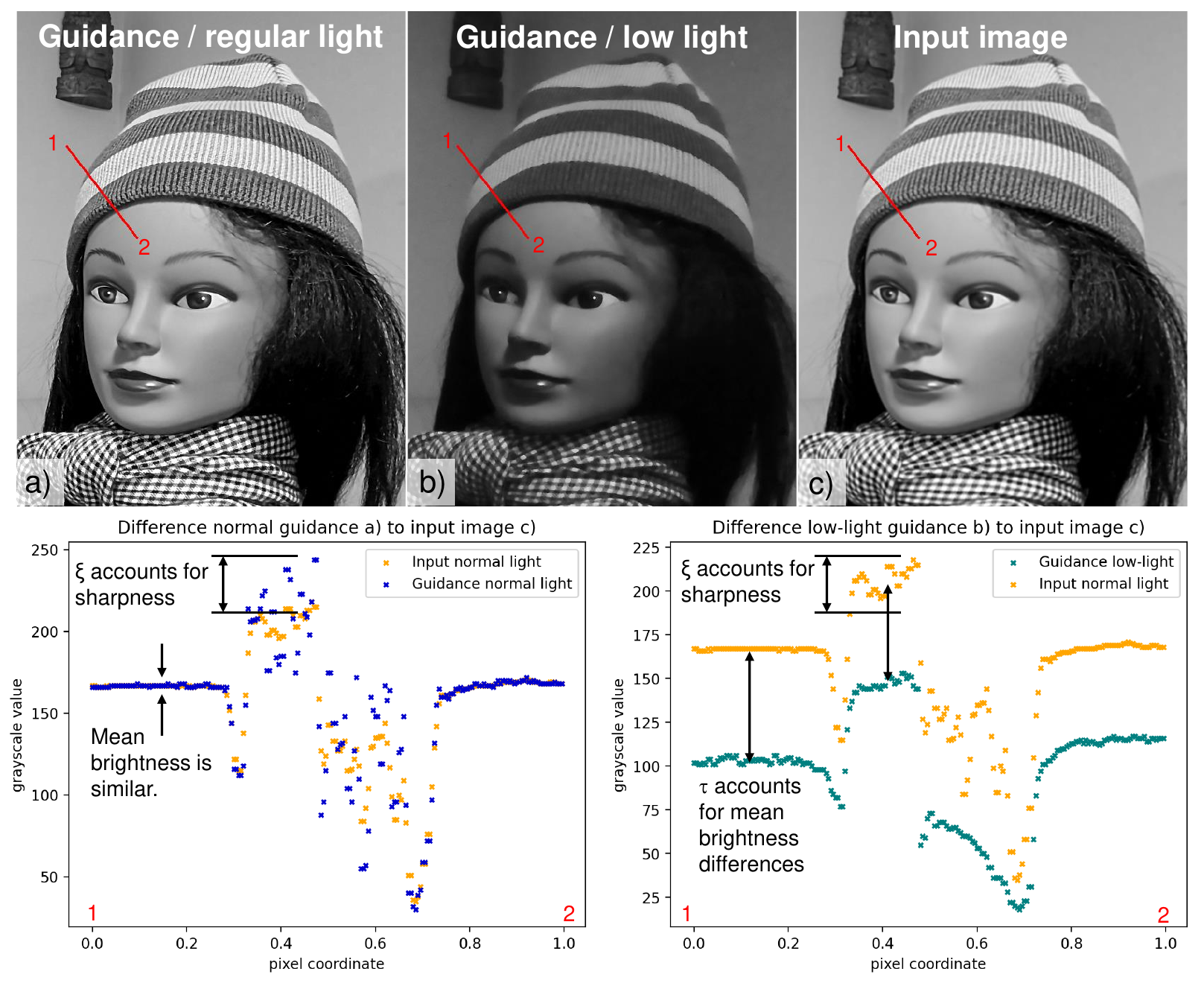}
\caption{Illustration of the offset difference between the original and low-light guidance image. The parameter $\tau$ accounts for this offset, while the parameter $\xi$ trains to restore sharpness (contrast gradient). Without account for the brightness offset between low-light and guidance image, $\xi$ trains for the offset difference but omits sharpness. }
\label{fig:offset}
\end{figure}

Figure~\ref{fig:offset} illustrates the problem and the effect of our solution. The figure shows a) a guidance image $G$ in regular light, b) a low-light image $G$, and c) an input image $I$ with reduced sharpness. The brightness values have been sampled along an edge (red line sampled across 1-2). 

The two charts show the gray-value distribution across the 1-2-line. The left chart depicts the situation with  $|I| \approx |G|$. % and the right chart the low-light situation with $|G| << |I|$. 
In the left chart, it is noticeable that the mean brightness for both images is the same. The input image is of low sharpness, which is primarily noticeable as lower contrast peaks, thus, lower sharpness. In this case, the parameter $\xi$ as used with AGF accounts for these peak contrast differences across edges; it increases slope and max/min values. During training, $\xi$ assumes a value per gradient that allows $I$ to adopt the magnitude of $G$ in high gradient areas, since the error $E$ between $I$ and $G$ is higher than the error in other areas. This allows the approach to increase edge slopes, yielding a sharp image.  

The right chart depicts the situation when the overall brightness is $|G| << |I|$. In this case, a mean delta between $I$ and $G$ yields a much higher loss decrease during gradient descent. Or in other words, it is easier for the training process to adjust for the brightness difference than for sharpness. The values for $\xi$ become appropriate to correct for brightness differences. However, they underperform when sharpening the edge slopes. Our results (Section 5 - ablation study) demonstrate this behavior. 

%%--------------------------------------------------------------

\subsubsection{Approach}
We use a brightness correction factor $\tau$ to account for the brightness difference in $G$ and $I$:

\begin{equation} \label{eq:tau}
    O_p = A_{k}(G_{p}+\tau_{p} + \xi_{p} )+B_{k}
\end{equation}

\noindent The equation is an extension of Eq.~\ref{eq:01}, adding $\tau$ as the new parameter. 

Mathematically, the goal of $\tau$ it to align the gradients of the input image $I$ with the gradients of the guidance image $G$ to restore optimal conditions for the linear transform (Eq. \ref{eq:calc_ab}). Assuming the relation between the gradients of the two images as (to simplify, in one direction)
\begin{equation}
 \frac{\text{d}G}{\text{d}x} >> \frac{\text{d}I}{\text{d}x}
\end{equation}

with $\text{d}G/\text{d}x$ and $\text{d}I/\text{d}x$, the gradients of the image, and $\text{d}G/\text{d}x >> \text{d}I/\text{d}x$ for the same content; and the image coordinates $x$. The guidance image is of lower brightness, but of higher resolution and more contrast than the enhanced image $I$, which yields higher gradients on average, thus, a sharpness difference. 

Our correction factor $\tau$ adds an additional component to align the gradient magnitude, such as:
\begin{equation}
\label{eq:inequality}
\frac{\text{d}G}{\text{d}x} = \frac{\text{d} (A\tau)}{\text{d}x} + \frac{\text{d} (AG + A\xi)}{\text{d}x}		
\end{equation}
with $\text{d}O = \text{d} (AG + A\xi)$ (we removed the coefficient $B$ for simplification). 

During training, $\tau$ accounts for the gradient indifference between the input and guidance image. AGF only uses $\xi$, causing the gradient imbalance. The magnitude of $\xi$ depends on gradient descent, thus, the second derivative of the gradient, or the first derivative of the loss function $\text{d}\xi = A (O-I)  \text{LoG}$ . One can see that in the case where the $LoG = 0$, the gradient is $0$, and $\xi$ remains low. In this case, the term $\xi$ does not contribute to satisfy Eq~\ref{eq:inequality}. 

Our approach augments $\frac{\text{d}(A\tau)}{\text{d}x}$. The magnitude of $\tau$ is driven by the brightness difference, $\text{d}\tau = A(O-I)$. It accounts for the general mean shift between $G$ and $I$, see the right chart Figure~\ref{fig:offset}. Once the mean-shift is corrected, $\xi$ can be trained to adjust for sharpness along edges.
 
\subsection{Training}

Practically, $\tau$ is determined for various brightness classes similar to \cite{PJ15} edge-class approach. We determine classes per pixel based on the brightness difference (mean difference) between $I$ and $G$. We discretize the range between brightness boundaries into N-classes. For each class, we compute a correction factor $\tau$. We optimize $\tau$ via gradient descent to identify the best parameter that accounts for the brightness in the training set. 

The training process in detail includes three steps:

First, we determine the brightness classes for each pixel of an image. We calculate the brightness difference per pixel $\delta = I_p - G_p$. Based on the $\delta$, the pixels of $G$ and $I$ are labeled by linearly splitting the min and max boundary range into $N$ classes (N=121, empirically determined) and assigning the label using $cb_i = \lfloor \frac{\delta}{\frac{max_I - min_I}{N}} \rfloor $ with $i$, the class index. 

Secondly, we assess the loss function for each class using: 
\begin{equation}\label{modified_loss_function}
    E(A_{k},B_{k}) = \sum_{p\in w_{k}} ((A_{k}G_{p}+B_{k} + \tau_{p} +\xi_{p})-I_{p})^{2}+\varepsilon A_{k}^{2}
\end{equation}

\noindent The process is equivalent to the loss as used in Eq.~\ref{eq:01}. Here, we add $\tau$ to the loss function and utilized the fact that brightness differences are overpowering and, thus, easier to correct and train than sharpness. When training for $\tau$, $\xi$ and $\sigma$ are set to constant zero.

\begin{figure}[ht]
\centering
\includegraphics[width =0.98\columnwidth]{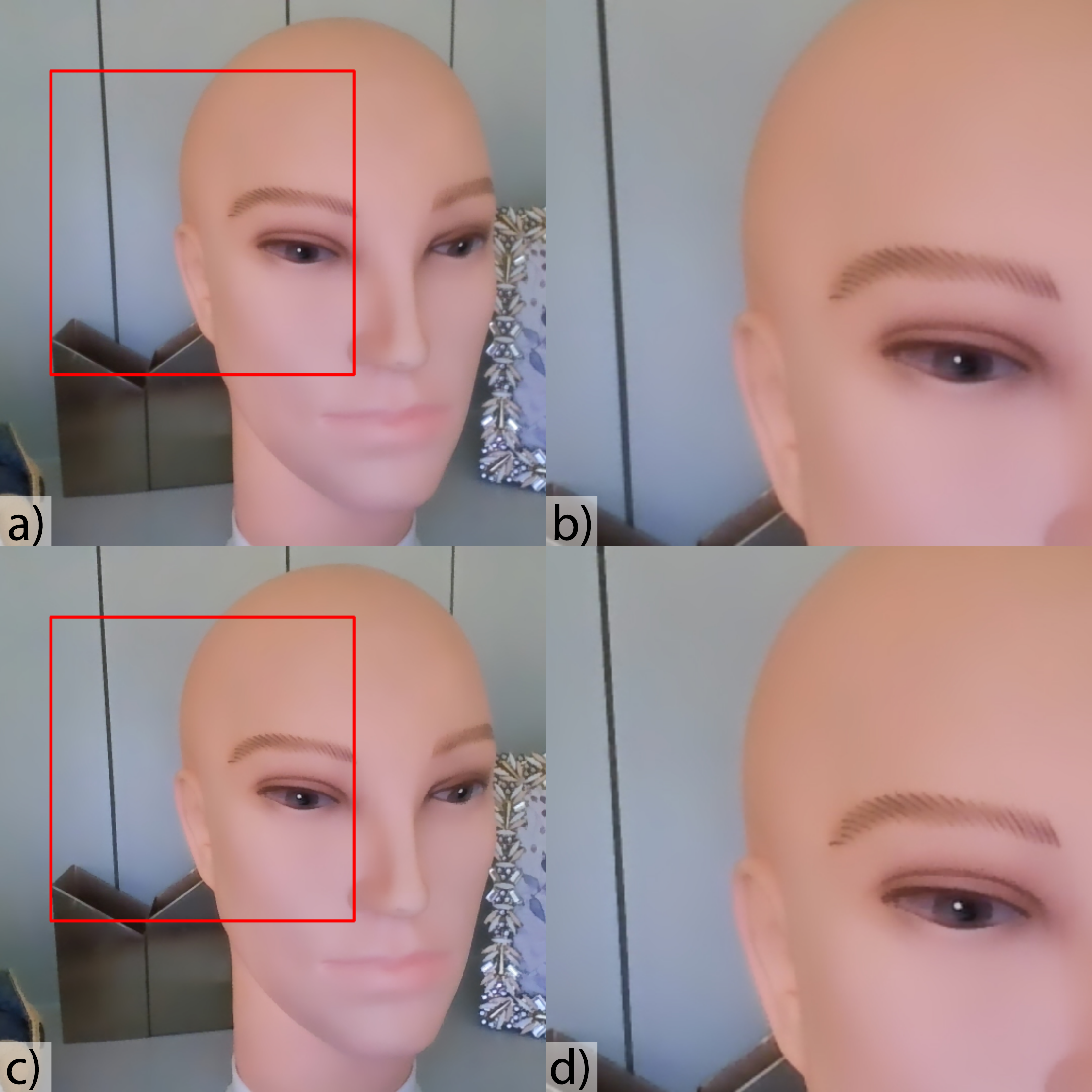}
\caption{Comparing AGF and AGU processing to demonstrate the impact of $\tau$, which accounts for the low brightness in the high-res guidance image. a) and b) is the AGU processed image and a magnified section of it with sharpness $s=14.68$, $\sigma^2=0.244$, $PSNR=37.88dB$ (compare to the enhanced image). Figures c) and d) are AGF processed, with $s=8.853$, $\sigma^2=0.281$, $PSNR=45.94dB$ }
\label{fig:Figure_tau}
\end{figure}

Using Eq.~\ref{eq:tau}, our results demonstrate increased sharpness along edges in comparison to the original approach. Figure~\ref{fig:Figure_tau} demonstrates the difference with and without using $\tau$; Figure~\ref{fig:Figure_tau}a) and c) respectively; c) depicts a result generated with AGU. One can see sharpness increase, especially around eyes and the mouth, where d) shows almost no enhancement in comparison to b). Other results demonstrate equivalent results, and we further describe them in Section~\ref{sec:results}.

\subsection{Scale-Agnostic Sharpness enhancement} \label{sec:scale}
Scale agnostic accounts for resolution differences between the guidance image $G$ and the input image $I$. One goal of guided upsampling methods is restoring the sharpness of the original image in the upscaled one. In our case, the guidance image $G$ is usually of higher resolution, whereas the input image $I$ is low resolution. During upsampling, our approach transfers the sharpness characteristics from $G$ to $I$ irrespective of typical blurring effects due to interpolation (Section~\ref{sec:sharpening}). Technically, we deliberately oversharpen the image by boosting the linear equation parameters in Eq~\ref{eq:01} such that the interpolation operation reconstructs the dominant sharpness features of $G$ in the output image $O^\uparrow$. We apply a targeted upsampling filter for this purpose, emphasizing strong edges that perceptually correspond to sharpness and omit uniform areas. As a result, AGU can overcome challenges of prior art and work with images of various scales.

\subsubsection{Problem}
Upsampling images is a standard task in image processing, and it is a well-known problem that common linear interpolation from a low resolution to a high resolution reduces the fidelity of the image \cite{KMDM:2007}. Especially high-frequency content is lost (\cite{PL:2015}), often representing details and textures. Various approaches have been suggested yielding better results than linear interpolation, such as non-linear interpolation \cite{JZ+:20, Dengwen:2010}, guided upsampling \cite{HS15, CA16, KMDM:2007}, and upsampling in frequency space \cite{PL:2015, RJD:2020}. All those approaches outperform linear interpolation, maintaining high fidelity when upsampling from a low-resolution prior. 

Our approach is based on AGF, which originally maintains the resolution of the image. However, our goal is to upsample the image. Although it is possible for us to upsample the image in a subsequent process using FGF \cite{HS15}, for example, this would increase computational complexity and contradict our low-performance goal. Instead, we aim for noise reduction and upsampling in a single step while maintaining sharpness.

\subsubsection{Approach} \label{sec:sharpening}

Our AGU approach is to embed upsampling as a part of the interpolation function using:
\begin{equation}
\label{eq:with_upsampling}
O^{\uparrow}_p = \mathcal{I}( {A_{k}} ) ( G_{p}^\uparrow + \xi_p + \tau_p ) + \mathcal{I}(B_{k})
\end{equation}
with $A_{k}$ and $B_{k}$ the pixel-wise linear coefficients and $\mathcal{I}$, an interpolation function. Here, $G_{p}^\uparrow$ and $O_{p}^\uparrow$ refer to the high-resolution guidance and output image, respectively. Note that we introduce the $^\uparrow$ notation in this section to indicate high-resolution data, where all other symbols indicate content of low-resolution. We omitted this notation earlier to simplify reading; however, we introduce it here to clearly distinguish between low-resolution and upsampled high-resolution content. 

%Challenge
To account for the larger resolution, we upsample the coefficients $A_{k}$ and $B_{k}$ using the interpolation function:
\begin{equation}\label{linear_coeff_interpolation}
\begin{split}
    A_{k}^\uparrow &= \mathcal{I}{(A_{k})}\\
         &= e_{xx} (a_{ii} A_{k} + b _{ij} A_{k}) + f_{yy}(c_{ji} A_{k} + d_{jj} A_{k}) + ecb_{c_i}\\
\end{split}
\end{equation}
\noindent Here, the coefficients $a_{ii}$, $b_{ij}$, $c_{ji}$, $d_{jj}$, $e_{xx}$, $f_{yy}$ are the bilinear interpolation parameters in the low-resolution image index with $i$, $j$; $x$, $y$ address the image coordinates in the high-resolution images. 

We introduce a class-based correction factor $ecb_{c_i}$  to prevent fidelity loss. The interpolation function is linear. Applying it without correction comes with the typical bi-linear interpolation disadvantage, such as sharpness reduction. The factor $ecb_{c_i}$ is class-based, with $c_i$, the class index, performing a targeted sharpening at strong edges. The classes distinguish between edges and uniform areas and ensure that strong edges are over-emphasized. Technically, we oversharpen dominant edges that contribute to perceptual sharpness to the level that applying Eq.~\ref{eq:with_upsampling} yields the correct target sharpness. 

The factor $ecb$ works as follows: The factor  adds or reduces the magnitude of $A$ and $B$ depending on the class, where the class depends on the gradient in the high-resolution image $G_p^{\uparrow}$. The higher the gradient, the higher the correction. Generally, we can distinguish two class ranges, one for high, positive gradients, and one for low, negative gradients, which can be expressed as:
\begin{equation}
ecb_{c_i} = \left\{ \begin{array}{ll}
f[LoG] &  if \frac{dG}{dx} > 0 \\  
-f[LoG] & \text{otherwise}
\end{array}
\right.
\end{equation}
\noindent Here, $f$ represents a look-up table with correction values, changing the linear interpolation result. Without the correction, the linear interpolation would lower an edge value to a mean, and the factor $ecb$ corrects the values of $A$ and $B$ to their original magnitude. For negative gradients, it corrects it to the original lower magnitude, and for high-magnitude gradients, it corrects it to the higher value. We need to point out that $ecb$ emphasizes strong edges in the image. The target value for $ecb$ for classes representing weak gradients or uniform areas is negligible.  

The class labels are determined using the guidance image $G^\uparrow$. We apply a bilateral filter to the image to emphasize edges and use the LoG-operator to identify relevant gradients (similar to  \cite{PJ15}). Using the LoG response, we linearly split the image into $N$ classes, determining a class label per pixel with:
\begin{equation}\label{L(G)_class}
c_{i} = \frac{ {LoG}(G^\uparrow)} { \lfloor range/N \rfloor}
\end{equation}
\noindent Although N is variable, we currently aim for 121 classes (empirically determined parameter).

We individually determine the actual magnitude of $f$ in a training pre-process for $A$ and $B$, a process explained in the next section.

\subsubsection{Training of $ecb$}

The values for $ecb_{c_i}$ are determined during a prior training using an iterative, gradient-descent approach optimizing the cost function $\mathcal{J}$:
\begin{equation} \label{eq:Itrain}
    \min \mathcal{J} = | {LoG}(G) - {LoG}(O^\uparrow) |
\end{equation}
\noindent with ${LoG}$, the Laplacian of Gaussian (LoG) operator. We assume that the response to a LoG-operator on an image represents the perceivable sharpness in this image, similar to \cite{PJ15}. Additionally, we apply a bilateral filter in advance to extract meaningful edges. 

\begin{figure}
\centering
\includegraphics[width = 0.98\columnwidth]{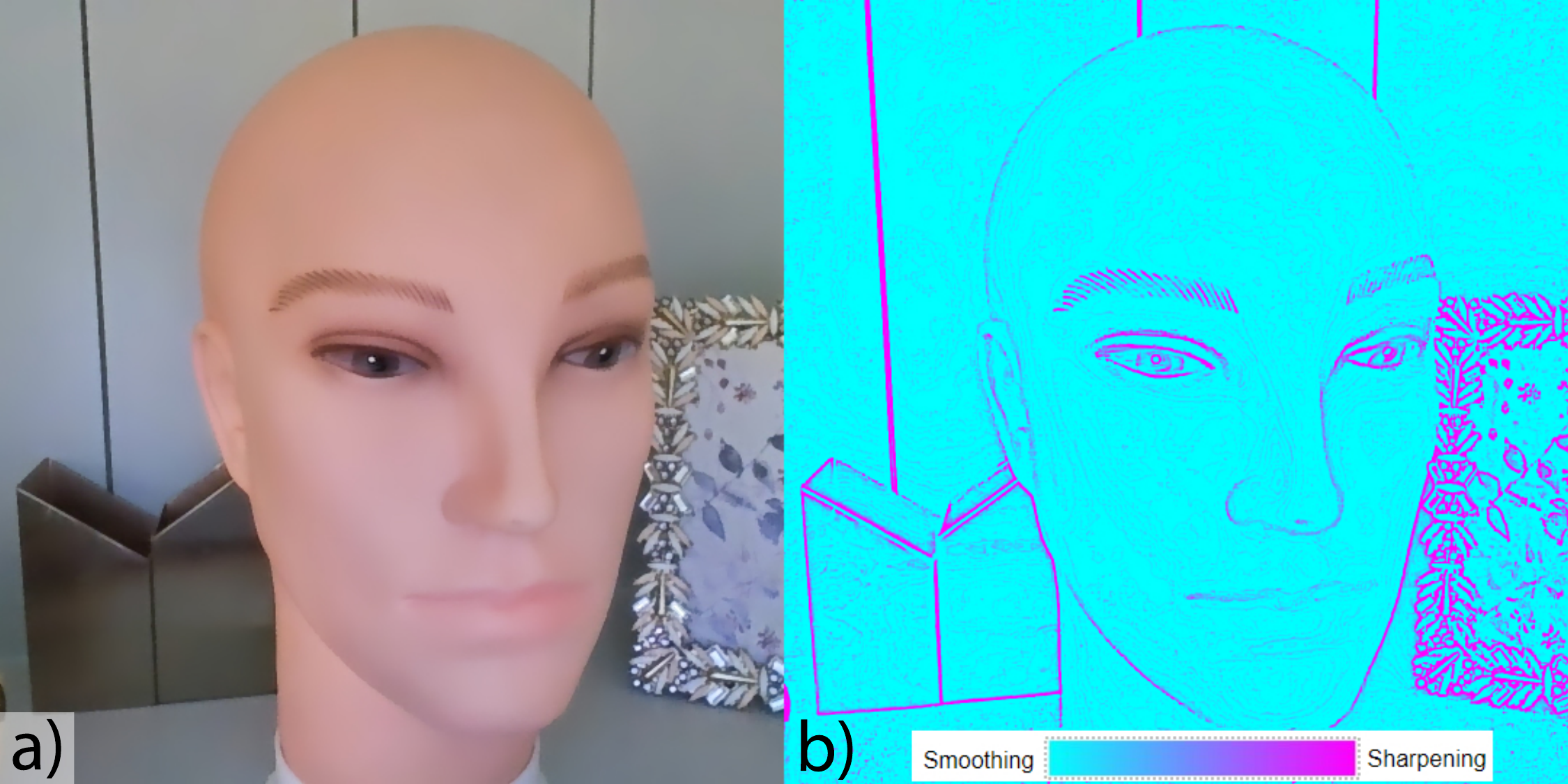}
\caption{A Laplacian of Gaussian Operator allows one to discriminate strong edges from uniform areas. We apply the filter on the high-resolution camera image (a) and generate labels (b) per pixel indicating areas to be smoothed or sharpened.}
\label{fig:LoG}
\end{figure}

To train $ecb$, we apply the LoG filter to $G^\uparrow$ and $O^\uparrow$, and minimize the difference between the image magnitude per class using:
\begin{equation}
\label{eq:sharp_train}
\begin{split}
   ecb_{c_i} = 
    \left\{ \begin{array}{ll}
    +const & {LoG}(G) < {LoG}(O^\uparrow)\\
    -const & otherwise
    \end{array} \right.
\end{split}
\end{equation}
We use a step-wise function adding or subtracting an additional class-based value until Eq.~\ref{eq:Itrain} reaches a minimum. 
The method as described yields a value per class. The class distribution allows us to distinguish between uniform areas and strong edges as shown in Figure~\ref{fig:LoG}. %The parameters for $ecb$ emphasize edges, as explained in detail in Sec~\ref{sec:class_labels}.

\subsubsection{Example Results}
% working for low-light images

\begin{figure} 
\centering
\includegraphics[width = 0.98\columnwidth]{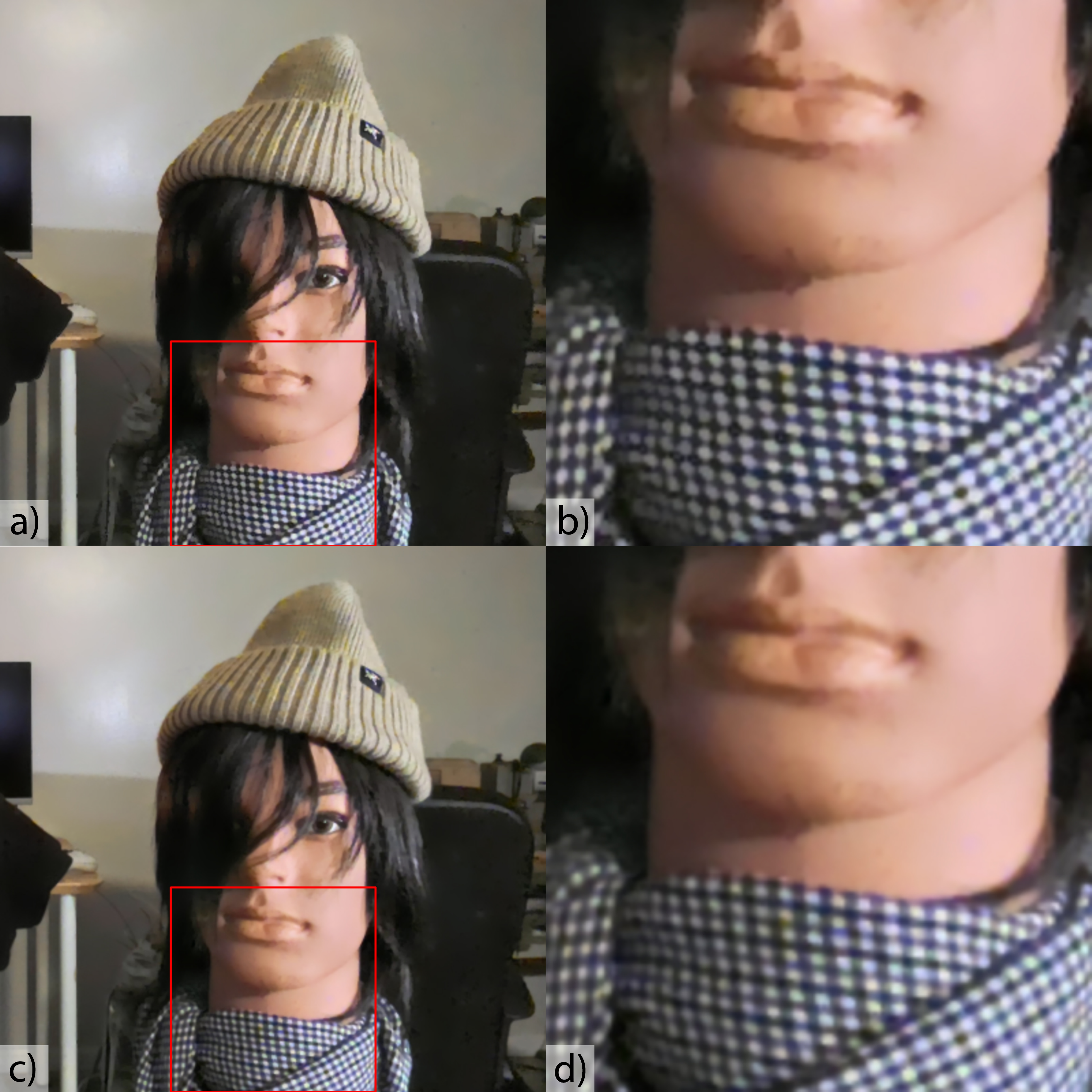}
\caption{The figure compares an upsampled image a-b) applying the correction $ecb$ to the upsampling interpolation, and c-d) without the correction factor. One can see that the correction factor emphasizes sharpness mostly along strong edges, which increases the overall sharpness perception.}
\label{fig:ecb}
\end{figure}

Figure~\ref{fig:ecb} demonstrates the effect of the correction factor $ecb$ applied class-based to an image processed with Eq.~\ref{eq:with_upsampling}. The comparison demonstrates the effectiveness of $ecb$ during upsampling. 

\noindent Comparing b) and d), one can see that the correction factor better emphasizes all edges around the eyes and the mouth.

\subsection{Algorithm}\label{algo}
This section outlines the algorithm for AGU. The linear coefficient A and B are calculated similar to the original AGF\cite{PJ15}(Note the comments in the pseudocode) everything else is our modification.The algorithm is as follow:

\begin{algorithm}
\caption{AGU pseudocode}\label{alg:cap}
\begin{algorithmic}[1]
\Require Lowlight Guidance Image $G^{\uparrow}$; kernel radius $k$;
\Statex image dimensions N;  no. of classes $n$ ; 
\Statex upsample factor $uf$
\Ensure Upsampled output image $O^{\uparrow}$
\State $Y^{\uparrow} \gets \textbf{conv}(G^{\uparrow})$
\State $Y_{max} \gets \min(Y,k)$
\State $Y_{min} \gets \max(Y,k)$
\State $c \gets \frac{ {LoG}(Y^\uparrow, k )} { \lfloor (Y_{max}-Y_{min})/n \rfloor}$
\State $cb \gets \frac{ ({I-Y})} { \lfloor 2max(RGB) /n \rfloor}$
\State $Y_{mean} \gets computeMean(Y,k)$ \Comment{Begin original AGF}
\State $Y_{corr} \gets computeMean(Y.*Y,k)$
\State $Y_{var} \gets Y_{corr} - Y_{mean}.*Y_{mean}$
\State $I_{mean} \gets computeMean(I,k)$
\State $IY_{corr} \gets computeMean(I.*Y,k)$
\State $IY_{cov} \gets IY_{corr} - I_{mean}.*Y_{mean}$
\State $\epsilon \gets \lambda(LUT_{\sigma}(c))^2$
\State $A \gets IY_{cov}./(Y_{var}+\epsilon)$
\State $B \gets I_{mean} - A.*Y_{mean}$
\State $A_{mean} \gets computeMean(A,k)$
\State $B_{mean} \gets computeMean(B,k)$
\State $\xi \gets LUT_{\xi}(c)$ \Comment{End original AGF}
\State $\tau \gets LUT_{\tau}(cb)$
\State $ecb \gets LUT_{ecb}(c)$
\State $A^{\uparrow} \gets upsample(A_{mean},ecb,uf)$ \Comment{Refer \ref{sec:scale} } 
\State $B^{\uparrow} \gets upsample(B_{mean},ecb,uf)$
\State $O^{\uparrow} \gets A^{\uparrow}(Y^{\uparrow}+\tau+\xi)+B^{\uparrow}$
\end{algorithmic}
\end{algorithm}

The above algorithm is applicable to a gray scale image. To apply the algorithm to color images we split the image into its component color channels and apply the algorithm per channel and then merge them to create the colored image.However, $c$ is always calculated on the gray scale image even if algorithm is being applied to color images. In line 6 $max(RGB)$ refers to the max color range an image can have, in our case 255.

The computational complexity of the algorithm is mainly driven by the upscale factor $uf$. Overall the image resolution and upscale factor $4\mathcal{O}({(N.uf)^2})$+$9\mathcal{O}({N^2})$ affect the runtime complexity.In cases where $uf$>1.5 the $uf$ dominates and the complexity can be computed as $\mathcal{O}({uf^2})$. 

\subsection{Learning Approach} \label{sec:train}
% Main message
% Sequential training
% Training sequence. 
Adaptive Guided Upsampling  relies on MLE to identify the optimal parameters. In total, four parameters need to be training:

\begin{enumerate}
\item $\tau$: a factor to account for brightness differences between the input and enhanced images. 
\item $\sigma$: the optimal smoothing factor for AGF.
\item $\xi$: the sharpening factor as described in AGF. 
\item $ecb$: sharpening factor that reflect the sharpness difference between the input and enhanced image.
\end{enumerate}

We train the four parameters sequentially on the results of the previous optimization steps. 
The first optimization process trains values for $\tau$ as described in Section~\ref{sec:brightness}. The process learns brightness shifts between the input $G$ and enhanced image $I$. Applied on the enhanced image, our adaptation of AGF can compensate brightness differences during training. 

The second optimization process focuses on $\sigma$ and trains the optimal variance with the effect of $\tau$. We refer to \cite{Kim:2004} for training details. The authors originally used exhaustive search, where our approach relies on gradient descent; results are comparable. 

Subsequently, we train for $\xi$ following the approach as described by Kim et al. (\cite{Kim:2004}). Also, for this parameter, we use gradient descent instead of exhaustive search. 

We deviated from the original approach (exhaustive search) since our use case introduced brightness differences and camera noise. Gradient descent on individual trained parameters renders the approach more practical for a larger variance of cameras and content. Also, training $\xi$ and $\sigma$ jointly with gradient descent negates their effect because the two parameters counteract each other. Although processing is guided by the $LoG$ classes, which would prevent adverse effects when properly trained. However, during training, values are ineffective in early training steps and can discourage the process from terminating.

Finally, we train $ecb$ using linear regression to account for the sharpness reduction due to upsampling. All other parameters are applied as trained during this final step.

\section{Experiments \& Results} \label{sec:results}

We conducted a set of experiments to verify the effectiveness of the proposed method. The experiments compare our approach to the related work and standard upsampling methods. %Goal is improved sharpness and noise reduction when working with low-light image enhancement and low-light guidance images. 

%The next sections introduce the datasets, metrics, and explain the results. 

\subsection{Dataset}
We use a publicly available dataset as well as Lenovo dataset. 
\begin{enumerate}
\item LOL dataset \cite{Wei:2018} comprises of 500 low-light and normal-light image pairs, divided into 485 training image pairs and 15 test pairs of $600\times400$ pixel resolution. The images cover indoor scenes and are subjected to sensor noise.  
\item Lenovo Low-light test set \cite{Lenovo:2024}, which comprises 100 low-light images at $1920 \times 1080$. The images have been recorded with a Lenovo ThinkPad X1 2-in-1 Gen 9, and show various indoor images all captured in low-light conditions. 
\end{enumerate}

The first dataset allows us to compare the approach to other solutions. The second one provides images more relevant to our use case.

\subsection{Methods and Metrics}  
We processed the dataset images with the suggested approach comparing the results with methods focusing on upsampling as well as on guided filters for image improvement. We compare our method to the Adaptive Guided Filter (AGF, \cite{PJ15}), the Fast Guided Filter (FGF, \cite{HS15}), and use a bilateral filter (BF) as a baseline. Since none of those filters' objective is upsampling, we also compare our results to Bilateral Guided Upsampling (BGU, \cite{CA16}, \cite{BGU16}) and a bilinear upsampling (BU). 

We do not upsample images processed with our method when comparing with AGF, FGF, and BF. All the methods maintain the resolution of the input image for the output image. We use a resolution of $960 \times 540$ in this case for $I$ and $O$ (see Figure~\ref{fig:Figure1}) and perform the operation on RGB images. We upsample the image to a resolution of $1920 \times 1080$ for image when comparing to AGU and BU. The input image $I$ is of size $960 \times 540$, and output image $O$ is of size $1920 \times 1080$, which restores the original camera resolution $C$. For the LOL dataset, we maintain the procedure, but the resolution changed to $600 \times 400$ for the low-resolution image and $1200 \times 800$ for the high-resolution image. 

To obtain results, we use several metrics:
\begin{enumerate}
\item Noise $\sigma^2$: we use the method as described in \cite{Immerkaer:1996}. It is a non-reference metric relying on the noise variance across the image to quantify the mean Gaussian noise. It allows us to assess the noise reduction per image. 
\item Sharpness $s$: we use the average Laplacian of Gaussian magnitude as a non-reference metric indicating the sharpness of an image. 
\item PSNR: We compute the PSNR value between $I$ and $O$ as an indication for the amount of change. A smaller PSNR indicates more difference with respect to image $I$, which is the goal. Note that sharpness and noise are still required to assess the quality of the outcome. 
\item SSIM: We use the Structural Similarity Index to evaluate the structural differences between $I$ and $O$.
\end{enumerate}

%To obtain results, we measure noise (\cite{Immerkaer:1996}) and sharpness (\cite{Kanjar:2013}) of each image and compare the sharpness increase and noise reduction between the four different methods. We also compare the result to a ground truth using PSNR. 

\subsection{Quantitative Results with Upsampling}
%Main sharpness and noise after upsampling
%high-res
%Objective: Demonstrate that AGU is better than bilinear upsampling, FGU.  
%Method: Comparison of AGU to other methods using sharpness, noise measure, PSNR & SSIM to target output

The objective of this experiment is to demonstrate the improvements AGU yields when upsampling the image. Goal is to maintain sharpness and reduce noise levels after upsampling. In this case, the images have been processed as illustrated in Figure~\ref{fig:Figure1}, using the proposed method AGU, and comparing it to Bilinear upsampling and BGU. Images have been upsampled from an input resolution $960 \times 540$ to an output of $1920 \times 1080$. 

Table~\ref{tab:sharpness} presents the sharpness results for bilinear upsampling, BGU, and AGU. The column lists the various methods, and the rows show the metrics. 

\begin{table}[h!]
\centering
\caption{Experimental results comparing the average performance for upsampled images. }
{\def\arraystretch{1.2}\tabcolsep=5pt
\begin{tabular}{ c | c | c| c| c  } 
%\hline
Metrics & input  & Bilinear & BGU   & AGU  \\
& (540p) & (1080p) & (1080p)	& (1080p)   \\
\hline
\textit{Lenovo dataset} &&&&\\
Sharpness  & 8.50 & 6.24  & 9.11  & 10.59\\
Noise      & 0.33 & 0.22  & 0.31  & 0.24 \\
PSNR       & n/a  & 47.44 & 41.71 & 37.13\\
SSIM       & n/a  & 0.82  & 0.82  & 0.83\\

\hline
\textit{LOL dataset} &&&&\\
Sharpness  & 19.901 & 12.33 & 19.01 & 19.85    \\
Noise & 0.773 & 0.407 & 0.782  & 0.791 \\
PSNR & 26.678 & 34.33 & 33.92 & 31.73 \\
SSIM & 0.261 & 0.752 & 0.813 & 0.892 \\
\hline
\end{tabular}}
\label{tab:sharpness}
\end{table}

The results demonstrate that AGU can maintain the best balance between sharpness and noise after upsampling the image from $540p$ to $1080p$. Compared to a Bilinear filter, AGU can significantly increase the sharpness and maintain noise, where the bilinear filter loses sharpness. Compared to BGU, AGU maintains a sharpness similar to BGU, however, noise reduction exceeds the results one can obtain with BGU. 

Note that the PSNR values is the lowest for AGU. Here, we compare the PSNR between the input and output image. The lower PSNR only means that AGU affects the image more significantly but does not allow to assess the quality of the image content since the input image is not the target. 

\subsection{Quantitative Results without Upsampling}
The objective is to compare the sharpness enhancement and noise reduction performance of AGU to state-of-the-art methods, which are a Bilateral Filter (BF), FGF, AGF, and AGU. Note that all three state-of-the-art techniques do not upsample the image. However, we like to compare the performance to those since AGU has been motivated by FGF and AGF. Images have been upsampled from an input resolution $960 \times 540$ to an output of $960 \times 540$. Note that AGU aims to "oversharpen" the image theoretically, but the effect is negated due to the upsampling interpolation. For this result, we maintained the resolution, which results in oversharpening. 

Table~\ref{tab:low_res} shows the results. The column shows the different images, and the rows the results for the Lenovo dataset and the LOL dataset. 

\begin{table}[h!]
\centering
\caption{Experimental results comparing the average performance for input and output images of same resolution. }
{\def\arraystretch{1.2}\tabcolsep=5pt
\begin{tabular}{ c | c | c| c| c  |c  } 
\hline
Metrics & input  & BF & FGF	& AGF & AGU-lr     \\
 & (540p) & (540p) & (540p)	& (540p) & (540p)   \\
\hline
\textit{Lenovo dataset} &&&&&\\
Sharpness  & 8.50 &	7.68 &	9.36	& 9.54  & 17.41	\\
Noise & 0.33 &	0.156 & 0.46 &	0.47	& 0.73	 \\
PSNR & n/a	&43.82 &	44.24	&42.83&	35.78\\
SSIM & n/a &	0.909&	0.82&	0.77	&0.82	\\

\hline
\textit{LOL dataset} &&&&&\\
Sharpness  & 19.901 & 17.162 & 17.243 & 18.183 & 27.81      \\
Noise & 0.773 & 0.574 & 0.661 & 0.676   & 0.691  \\
PSNR & 26.678 & 38.21 & 34.32 & 34.45   & 31.27  \\
SSIM & 0.261 & 0.937 & 0.831 & 0.646  & 0.821  \\
\hline
\end{tabular}}
\label{tab:low_res}
\end{table}

As expected, the results indicate that AGU increases the sharpness tremendously, however, which results in image oversharpening in this case. We like to repeat that this is not the goal, and that a resolution change along with a bilinear interpolation negates this oversharpening effect. The Bilateral filter's performance can reduce noise significantly, but it also comes with the lowest sharpness. The images start to show blur along edges, and loss of details. Although it outperforms other standard filters, it is below guided solutions. The results for FGF and AGF are comparable. Note that the noise reduction capability of FGF is limited since the guidance image is a low-light image with a significant higher noise level than the input image. In this case, FGF also transferred the noise characteristics back to the input image. The performance of AGF is a result of failed training. Although AGF performs properly when the brightness delta between the input and guidance image is low, we were not able to yield any significant improvements from AGF in our use case. 

For the LOL dataset, AGU also performs as expected and quantitative results show and sharpness improvement for upsampled images while the noise is on an acceptable low level. All results are comparable to the Lenovo dataset with two differences. AGU tends to oversharpen the images slightly, but noticeably. Pixel artifacts along strong edges become noticeable when zooming in into details. Note that oversharpening is likely a result of the dataset and the artificial brightness reduction. Also, the images are sharper than typical laptop camera images. 

The results BGU provides are comparable to AGU for sharpness improvement. However, the quantitative results demonstrate that it lacks noise reduction capabilities when applied to the Lenovo dataset. The results are equivalent when compared to the LOL dataset. We need to point out that the LOL dataset has been captured using a DSLR camera and brightness reduction is artificial. The images come with less noise from the beginning.

\subsection{Qualitative Results}

Figures~\ref{fig:LLNetresults} and \ref{fig:LOLresults} demonstrate qualitative results obtained from the Lenovo dataset and the LOL dataset respectively. Note that the images are of different resolutions since they depict upsampled results at a resolution of $1920 \times 1080$ as well as $960 \times 540$. Advantages of AGU are noticeable when observing edges and details. Edges are better preserved, which contributes to the overall perception of sharpness. 

\begin{figure}[ht]
\begin{center}
\includegraphics[width = 0.99\columnwidth]{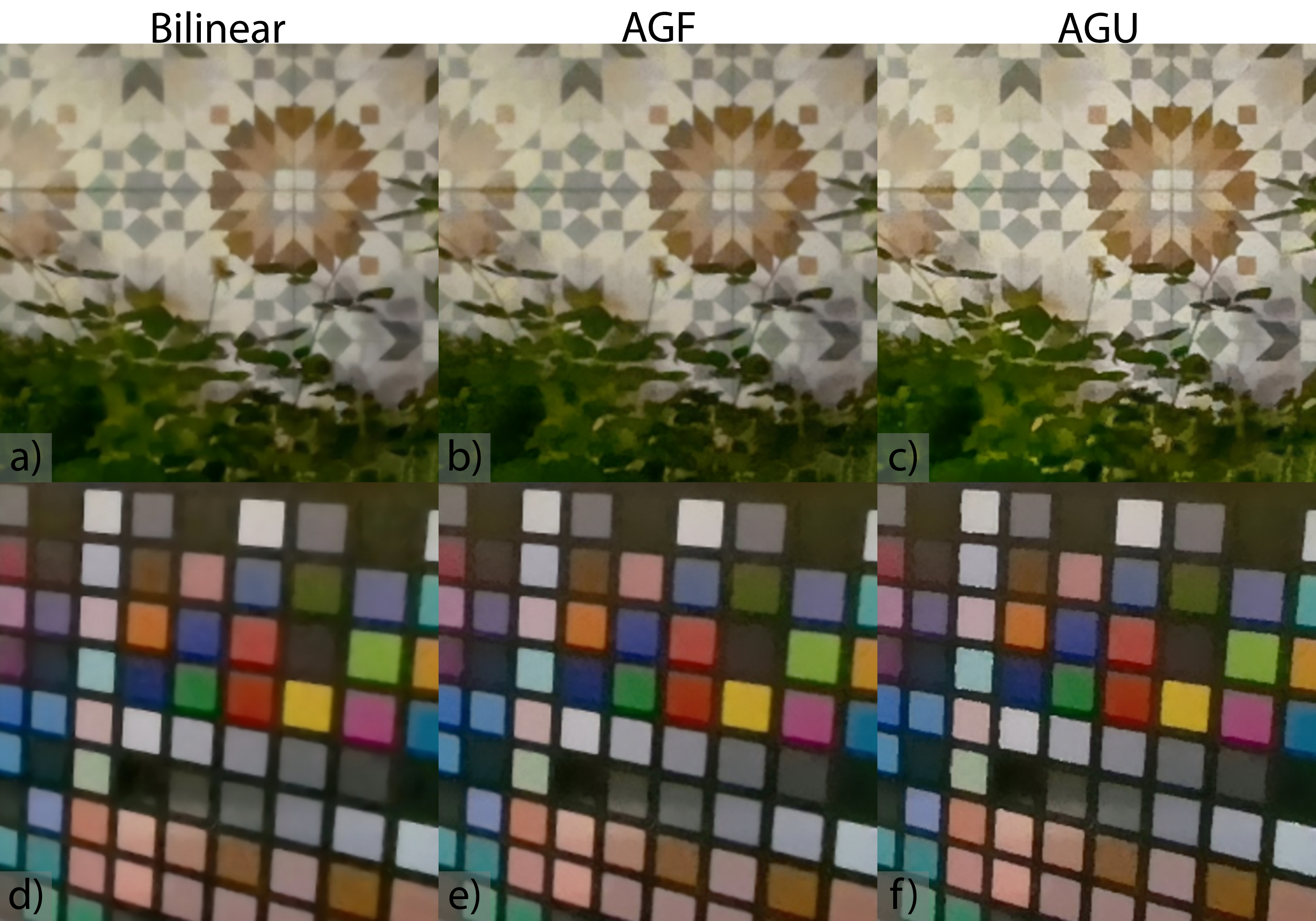} 
\caption{Detailed views after applying a,d), a bilinear filter, c,e) AGU, and d,f) AGF. Note that the AGU image has been resizes so that figure size is equal for all views.  }
\label{fig:sharpness}
\end{center}
\end{figure}

Figure~\ref{fig:sharpness} shows some details from selected figures comparing bilinear upsampling with AGF, and AGU. AGU yields sharper edges than AGF or the bilinear filter. AGU emphasizes the sharp edges in images and improves them significantly in comparison to all the other edges. This is especially noticeable at defined edges, for instance, at the wallpaper in Figure~\ref{fig:sharpness}a-c) or the checkerboard edges in d-e).

\begin{figure}[ht]
\begin{center}
\includegraphics[width = 0.99\columnwidth]{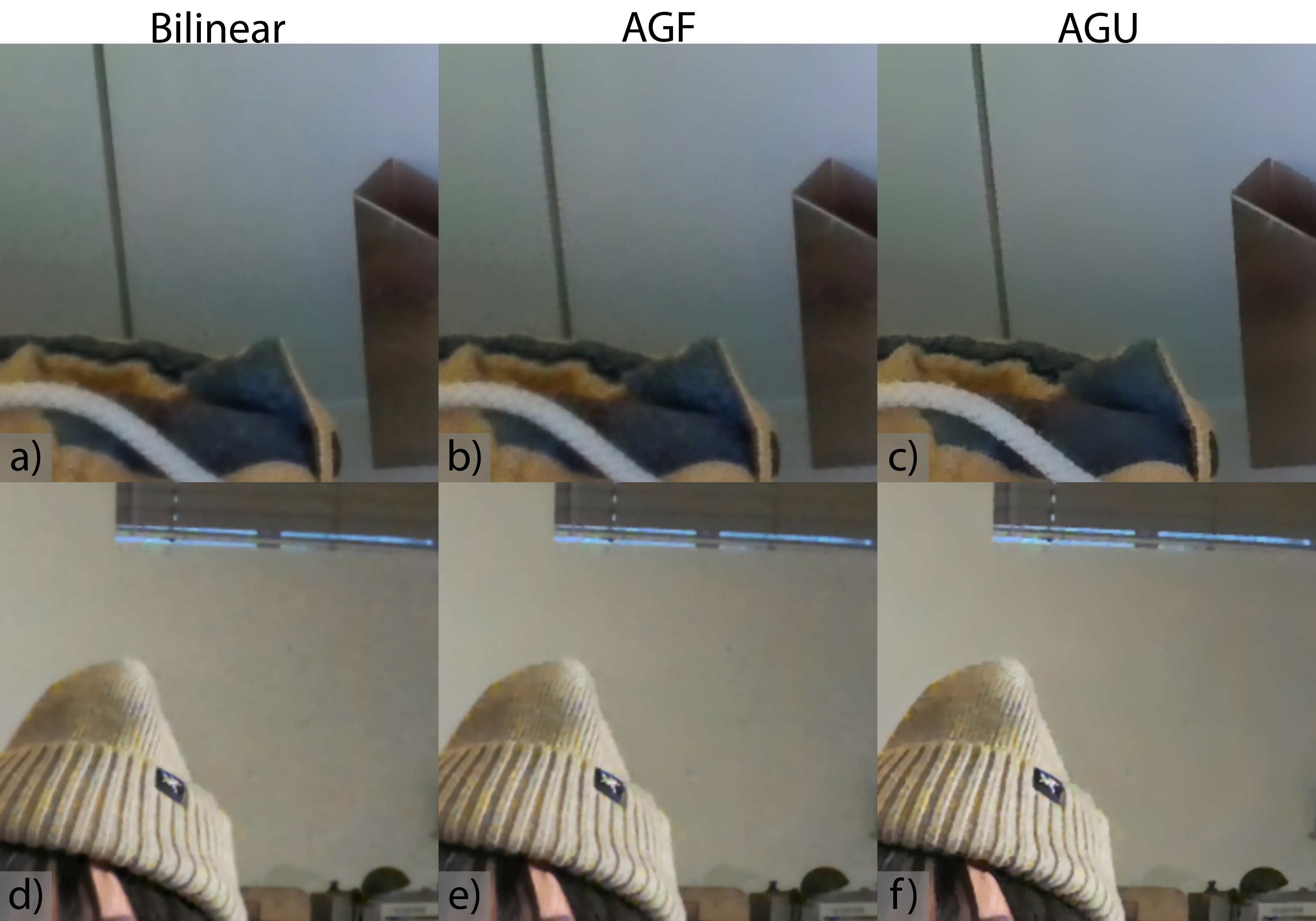} % 0.99
\caption{Detailed views after applying a,d), a bilinear filter, b,e) AGU, and c,f) AGF. Note that the AGU image has been resizes so that figure size is equal for all views.  }
\label{fig:noise}
\end{center}
\end{figure}

In addition, Figure~\ref{fig:noise} shows the noise reduction performance of AGU and compares it to a Bilinear Filter and AGF. AGU emphasizes uniform areas and yields peak noise reduction performance in large, homogeneous regions of an image. AGU is tuned in this manner to provide optimal performance in typical video conference use cases. The results in the images demonstrate a clear noise reduction from a) to c) and from d) to f) 

\subsection{Upsampling Ablation Study}

% Needs to show the impact of ecb on sharpness and noise reduction
We conducted a study to analyze the effect of our upsampling method, especially the upsampling correction factor $ecb$ and its impact on sharpness and noise. The parameter is trained for various sharpness classes and the analysis aims to demonstrate its effectiveness: maintain sharpness despite of the interpolation and to keep noise level at the target. Too much noise reduction blurs details and textures. We compare the results of upsampling with and without the correction factor $ecb$ for this purpose and measured sharpness and noise.

\begin{figure}[t]
\begin{center}
\includegraphics[width = 0.99\columnwidth]{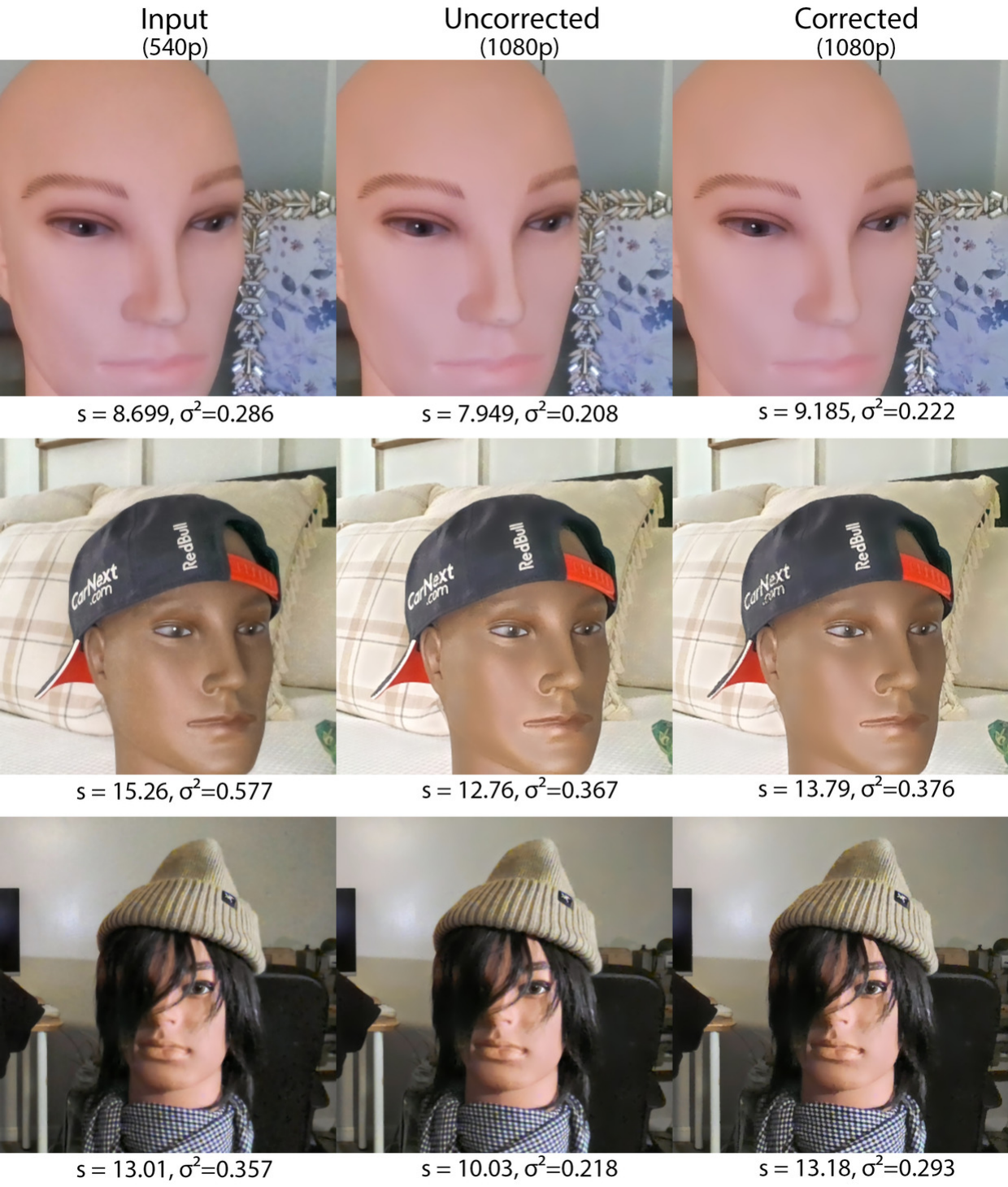} %0.99
\vspace*{-16mm}
\caption{Results comparing the effect of our interpolation correction factor to the input. The two right columns show the result yielded with and without the parameter $ecb$.  }
\label{fig:upsampling_ablation}
\end{center}
\end{figure}

Figure~\ref{fig:upsampling_ablation} compares results. Here, input is the enhanced image $I$ at a resolution of $960 \times 540$. The term 'corrected' refers to applying Eq.~\ref{linear_coeff_interpolation} for upsampling with $ecb$ correction, where 'uncorrected' omits this parameter. In this case, upsampling uses standard bilinear interpolation. The target resolution for upsampling is $1920 \times 1080$.

The quantitative results (measurements for sharpness $s$, noise $\sigma^2$) show a noticeable increase in sharpness when $ecb$ is applied. Similar to what we showed previously, strong edges are emphasized despite upsampling the image to a higher resolution. The sharpness is especially observable at areas with details, around the eyes, eye browns, text, or any other details in the image. In addition, the noise level remains low with only insignificant effect on details. This result extends to the entire Lenovo dataset with an average sharpness $s=10.59$, noise $\sigma^2=0.34$, $PSNR=37.13dB$, and $SSIM=0.83$. This sums up to a sharpness gain of $60\%$. Note that this sharpness gain is selective, improving edges and details. 

\begin{figure}[ht]
\begin{center}
\includegraphics[width = 0.99\columnwidth]{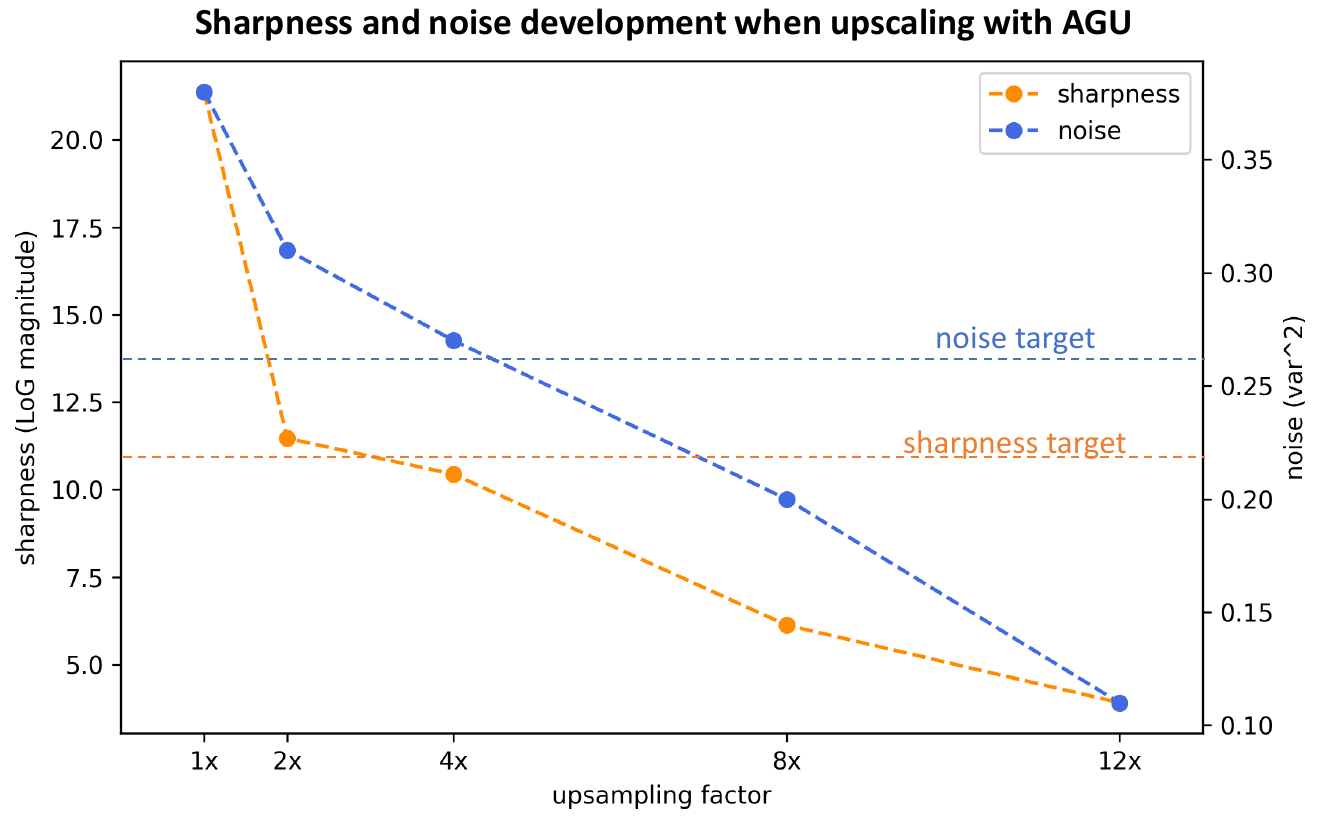} % 0.99
\caption{Sharpening and noise reduction effectiveness of AGU for various upsampling resolutions. Base resolution is 960$\times$540.  }
\label{fig:upsampling_scale_effectiveness}
\end{center}
\end{figure}

Figure~\ref{fig:upsampling_scale_effectiveness} shows the effectiveness for various upsampling resolutions, with the x-axis plotting the upsampling factor in comparison to the base resolution $960 \times 540$. The y-axis plots the average sharpness and the noise for the Lenovo dataset.  

%\newpage

The quantitative results demonstrate that the suggested method can maintain sharpness and noise up to an upscaling factor of $4\times$. Beyond this,  sharpness and noise decline noticeably. In addition, the noise level is also on an acceptable target level. For upscaling factors beyond $4\times$, the high noise reduction is a result of blurring due to too much upsampling. Note that these are the average quantitative results from the entire Lenovo dataset. Practically, the image up to a scale factor of $2\times$ are usable. Using a higher scale factor such as $4\times$ starts to show image degradation and blur.

\subsection{Guidance Image Ablation Study}
This analysis aims to demonstrate the effectiveness of the brightness correction factor $\tau$. It accounts for brightness differences between guidance image $G$ and input image $I$. The study compares results to corresponding target images with and without $\tau$. The factor ensures that the parameter $\xi$ accounts for sharpness differences. Otherwise, it trains to adjust the mean brightness delta between $G$ and $I$. The study demonstrates its effectiveness.  

\begin{figure}[ht]
\begin{center}
\includegraphics[width = 0.99\columnwidth]{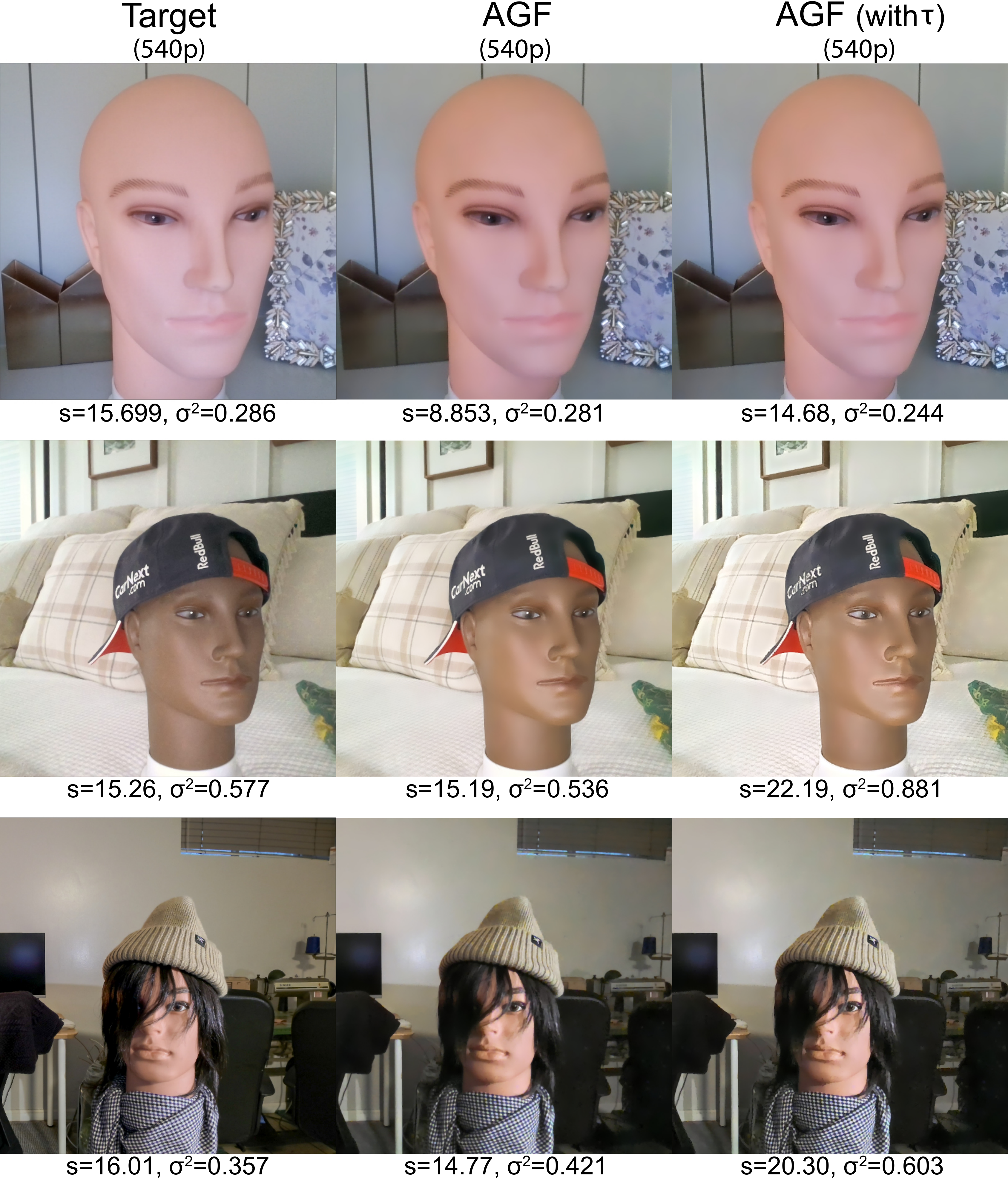} %0.99
\caption{Results comparing the effect of our brightness correction factor. The two right columns show the result yielded with and without the parameter $tau$.  }
\label{fig:brightness_ablation}
\end{center}
\end{figure}

Figure~\ref{fig:brightness_ablation} compares the result with an without $\tau$ to the target image. Without $\tau$ (column AGF), the sharpness increase is marginal compared to the input image (see in Figure~\ref{fig:upsampling_ablation}, left column). However, there is a noticeable discrepancy to the target image. AGU, and $\tau$, allows to restore sharpness, as one can observe in the right image. Quantitative results underpin this observation with an average $s=19.13$ for AGU in comparison to $17.41$ without $\tau$, $PSNR=34.34dB$ to $43.99dB$. The difference is noticeable when observing details and areas such as eyes, eye browns, and any object with a detailed texture. Note that we can observe a noticeable variance in the sharpness results, which is content-driven. AGU favors strong edges in an input image and is more effective in their presence. 

Note that AGU deliberately "oversharpens" the image in the case where input resolution is equal to the output resolution. Oversharpening is not noticeable in our typical upsampling use case, since the upsampling interpolation negates oversharpening effects. 

%%%%%%%%%%%%%%%%%%%%%%%%%%%%%%%%%%%%%%%%%%%%%%%%%%%%%%%%%%%%%%%%%%%%%%%%
% Main result images

\begin{figure*}[h!]
\begin{center}
\includegraphics[width =1.95\columnwidth]{Figures/Figure_9.pdf} %1.95
\caption{The following images show sample results starting with a), the low-light camera image, b) the output image of the low-light enhancer, which is equivalent to the input image for AGU. The following images show results processed with different methods c) bilinear upsampling, d) AGF, and e) our AGU. }
\label{fig:LLNetresults}
\end{center}
\end{figure*}

\begin{figure*}[h!]
\begin{center}
\includegraphics[width = 1.95\columnwidth]{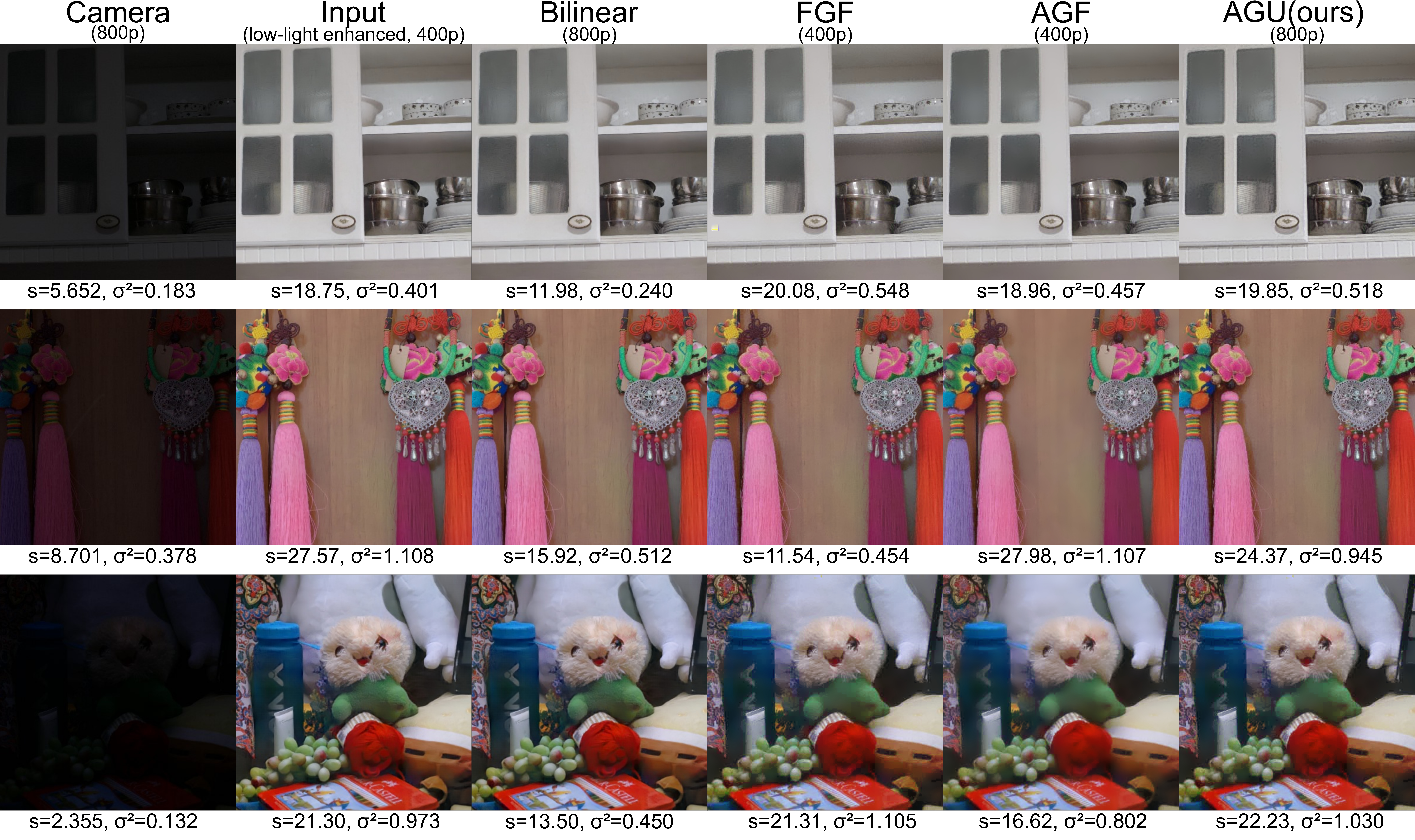} %1.95
\caption{The following images show sample results starting with a), the low-light camera image, b) the output image of the low-light enhancer, which is equivalent to the input image for AGU. The following images show results processed with different methods c) bilinear upsampling, d) AGF, and e) our AGU. }
\label{fig:LOLresults}
\end{center}
\end{figure*}

%%%%%%%%%%%%%%%%%%%%%%%%%%%%%%%%%%%%%%%%%%%%%%%%%%%%%%%%%%%%%%%%%%%%%%%%
% Main result images
\subsection{Runtime measurements}
We are required to process all camera image in camera frame time/real time. In low-light conditions, the camera runs at a typical frame rate of 15fps, equivalent to 66.66ms per frame. Therefore, the low-light enhancer and AGU cannot exceed this limit. This requires AGU to be substantially faster than the low-light enhancer process. Given our analysis, we expect the runtime for AGU to grow quadratically based on the upsampling factor. As demonstrated in Sec~\ref{algo}, an upsampling factor $>$1.5 causes a steeper runtime increase compared to increasing the image resolution.

The following Table \ref{tab:runtime} shows the runtime performance for Algorithm~\ref{algo} and the low-light enhancer (LLNet). The table lists the runtime for AGU and the low-light video enhancer respectively. We conduct the measurement for one base resolution, which we upsample to various output resolutions. The values are the mean out of 100 samples recorded per output resolution. 

\begin{table}[t]
\centering
\caption{Experimental results comparing the average runtime for upsampled images using different upsampling factors. }
{\def\arraystretch{1.2}\tabcolsep=5pt
\begin{tabular}{ c | c c c c } 
\hline
Base  & Output res & Upsampling   & Runtime & Runtime\\
res &  &  factor  & AGU ($\mu$s) & LLNet ($\mu s$)\\
\hline
256x& n/a & n/a &	n/a & 52100  \\ 
512&848x480 & 3.1 &	12.9 & 63800 \\
&960x540 & 3.9 &	14.1  & 66400 \\
&1280x720 & 7.0	& 14.8 & 106800 \\
&1920x1080 & 15.8 & 15.9 & 230400 \\
\hline
\end{tabular}}
\label{tab:runtime}
\end{table}

The results demonstrate that AGU does not add significantly to the overall runtime and that the solution can maintain camera frame time.

All measurements have been conducted on a ThinkPad X1 Carbon Gen 12, Intel Core Ultra 7 (Intel MTL), 155H, device with 16GB memory. Algorithm has been parallelized using OpenCL to meet runtime requirements.

\subsection{Discussion}
% low-light guidance
The results clearly demonstrate the advantages of AGU when working with low-light guidance images. The quantitative results indicate that sharpness improvement and noise reduction are superior to state-of-the-art guidance-based methods when working with guidance images of lower brightness. Figures~\ref{fig:LLNetresults} and \ref{fig:LOLresults} show improved sharpness, especially along strong edges,  compared to AGF. The method also exceeds the results of standard methods for noise reduction, such as the bilateral filter. Although the bilateral filter suppresses noise drastically, edge preservation has limits. 

% upscaling
Blur due to upscaling caused by any interpolation can be mitigated with our trained scalar correction factor. The results demonstrate that the linear "oversharpening" correction factor can account for the sharpness reduction caused by interpolation. AGU can successfully learn the difference before using ground truth and process images. The results indicate that focusing on strong edges is sufficient to restore most of the perceivable sharpness. Our method corrects uniform areas only gently; the results show that this is a feasible approach since those areas have a limited impact on the overall visual quality. Since the concept of our method (sub-sampling and correction) has been used for similar upsampling approaches (\cite{HS15} and \cite{CA16}), our results demonstrate that this approach works well for a class-based approach combining sharpness improvement and noise reduction.  

%%%%%%%%%%%%%%%%Runtime discussion
Our runtime experiments clearly show that AGU can run in real-time. Our results in Table~ref{tab:runtime} demonstrate that AGU requires $14.42{\mu}s$ on average for various resolutions. It also meets our real-time requirements when running subsequently to the low-light video enhancer. With the highest upscale factor, AGU is 14.5$\times$ faster than the neural network, which demonstrates the practicality of using a conventional algorithm alongside a neural network.

% Limitation: resolution delta.
Although the overall outcome meets our expectations, upscaling is limited. We noticed that sharpness improvement declines when using upscaling factors beyond $2\times$. Beyond this factor, the visual sharpness performance declines despite good sharpness data; blur starts to be evident in images with a scale factor of $4 \times$ or higher. This is likely due to the scalar correction factor we use. The scalar correction is unable to account for a wider interpolation distance. As a result, the image becomes undersampled and grainy. Thus, our method lacks scale-invariance when it needs to upsample an image to various resolutions. 

% ---------------------------------------------------
We must point out that the training dataset needs to be very well-curated. All target image attributes need to be harmonized, close to being similar. Training is sensitive to outliers. Any outlier, for example, one target image with too much noise, biases the training results, and thus, the correction factors. In this case, the correction factor variance per class is too wide, and the mean value does not represent the intended sharpness class well. The entire method underperforms noticeably in this case. This issue becomes especially problematic when using a larger training set. Thus, limiting the training data to ten or fewer samples is imperative. 

Also, the method as described focuses on low-light image enhancement. Guidance and input images are of different brightness and resolution. We rely on the high-resolution, low-light camera images as guidance images. Downscaling the camera image is essential for fast processing. The subsequent low-light enhancer changes the brightness significantly. Without these two conditions, our method yields results similar to AGF for sharpness improvement/noise reduction and BGU for upsampling. AGF primarily underperforms in this use case since the training of values for $\xi$ per class only corrects for brightness in this use case. 

We can also observe a significant difference between images captured with the target camera (Lenovo data) and images with artificial brightness reduction or noise. Although artificial noise and other image synthetic artifacts are feasible solutions for noise reduction algorithm evaluation, they still lack the dynamic of captured images and results do not reflect real-world outcomes. Especially in low light, images suffer from noise degradations, 3A camera problems, and darkness. Simulating these effects is challenging. The data (Table~\ref{tab:sharpness}). along with the example images, show that our method performs as expected for captured data. Using LOL data, we see improvements, but there is a tendency to oversharpen as well as blur. 

One of the remaining challenges is to decide the balance between noise reduction and sharpness improvement customers would accept when experiencing camera apps. In other words, when is noise not disturbing or edges sufficiently sharp. Our approach to this question is to use an image captured in bright conditions as a reference. We replicate bright-image characteristics in low-light images, assuming that customers like our bright-light camera images. We have indications that this target is acceptable to customers; however, we plan to confirm this with future user studies.

\section{Conclusion \& Outlook}\label{sec:conclusion}

In summary, we presented Adaptive Guided Upsampling: A method performing noise reduction, sharpness improvement, and upsampling simultaneously. Compared to state-of-the-art methods, it contributes a solution for brightness differences between the guidance and the input image as well as resolution differences between the input and output image. We demonstrate that a  correction accounts for brightness differences during training. We also show the feasibility of a constant interpolation correction to maintain image characteristics during bilinear upsampling. Thus, we conclude that our method performs as expected for our low-light enhancement use case. It can successfully upsample the image, without sacrificing sharpness or significantly increasing noise. Thus, it meets our expectations and requirements. It is an appropriate solution for video conference solutions using a low-light enhancer.

Future work will focus on a non-linear upsampling correction. As demonstrated, the linear correction can successfully maintain image properties up to an upscaling factor of two. We assume that a non-linear solution will further increase the upscaling factor. Also, our next steps will focus on improved noise reduction and keeping sharpness and details at an appropriate target level. Although noise reduction performs as expected, there is a limit when details start to become blurry. We aim to improve the regularization of training for this purpose since it contributes most to the loss of details. A different global regularization for smoothing should maintain details. We target an improved edge classification method. The Laplacian of Gaussian approach performs appropriately to meet the goal of this contribution; we intend to study edge detection methods and focus on better discriminating strong edges from other content. At the same time, we prepare to analyze whether a smoother class distribution between the two boundaries allows us a better balance between details and significant edges. Finally, we plan to analyze whether the method can be used to generate training data for a neural network and whether an end-to-end model can yield the same outcome.

%\bibliographystyle{ieeetr}
%\bibliography{reference}

\end{document}